%% file: main.tex
\documentclass{article}

\PassOptionsToPackage{numbers, compress}{natbib}

\usepackage[preprint]{neurips_data_2024}

\usepackage[utf8]{inputenc} 
\usepackage[T1]{fontenc}    
\usepackage{url}            
\usepackage{booktabs}       
\usepackage{amsfonts}       
\usepackage{nicefrac}       
\usepackage{microtype}      
\usepackage{xcolor}         

\input{math_commands.tex}

\definecolor{Gray}{gray}{0.9}
\definecolor{citecolor}{HTML}{2980b9}
\definecolor{linkcolor}{HTML}{c0392b}

\usepackage[pagebackref=true,breaklinks=true,colorlinks,bookmarks=false,citecolor=citecolor,linkcolor=linkcolor]{hyperref}

\usepackage{url}
\usepackage{graphicx}
\usepackage{enumitem}
\usepackage{booktabs}  
\usepackage{multirow}
\usepackage{wrapfig}

\usepackage{mdframed}
\newcommand{\sectioncolor}{violet}

\definecolor{Gray}{gray}{0.9}

\newcommand{\names}{\textsc{MultiIoT}}

\title{\names: Benchmarking Machine Learning\\for the Internet of Things}

\author{%
  Shentong Mo, Louis-Philippe Morency, Ruslan Salakhutdinov, Paul Pu Liang\\
  Carnegie Mellon University\\
  \texttt{shentongmo@gmail.com}
}

\begin{document}

\maketitle

\vspace{-4mm}
\begin{abstract}
The next generation of machine learning systems must be adept at perceiving and interacting with the physical world through a diverse array of sensory channels. Commonly referred to as the `Internet of Things (IoT)' ecosystem, sensory data from motion, thermal, geolocation, depth, wireless signals, video, and audio are increasingly used to model the states of physical environments and the humans inside them. Despite the potential for understanding human wellbeing, controlling physical devices, and interconnecting smart cities, the community has seen limited benchmarks for building machine learning systems for IoT. Existing efforts are often specialized to a single sensory modality or prediction task, which makes it difficult to study and train large-scale models across many IoT sensors and tasks. To accelerate the development of new machine learning technologies for IoT, this paper proposes \names, the most expansive and unified IoT benchmark to date, encompassing over 1.15 million samples from 12 modalities and 8 real-world tasks.
\names\ introduces unique challenges involving (1) generalizable learning from many sensory modalities, (2) multimodal interactions across long temporal ranges, (3) extreme heterogeneity due to unique structure and noise topologies in real-world sensors, and (4) complexity during training and inference. We evaluate a comprehensive set of models on \names, including modality and task-specific methods, multisensory and multitask supervised models, and large multisensory foundation models. Our results highlight opportunities for ML to make a significant impact in IoT, but many challenges in scalable learning from heterogeneous, long-range, and imperfect sensory modalities still persist. We release all code and data at the repository~\footnote{\url{https://github.com/Multi-IoT/MultiIoT}} to accelerate future research in machine learning for IoT.
\end{abstract}

\input{SECTIONS/10_Introduction/title_image}

\vspace{-4mm}
\section{Introduction}\label{sec:intro}
\vspace{-2mm}

The next generation of machine learning systems will need to understand and interact with the physical world through physical sensors. This interconnection of sensors is typically called the Internet of Things (IoT) ecosystem, which includes motion, thermal, geolocation, depth, wireless signals, pose, video, and audio to model the states of physical environments and the humans inside them~\cite{atzori2010internet,li2015internet,rose2015internet}. These sensing technologies have had great impact in recognizing human physical activities to inform us of our daily physical wellness~\citep{liang2021learning,qi2015survey,yuehong2016internet}; navigating self-driving cars and efficiently connecting them with transportation grids~\citep{javaid2018smart,khayyam2020artificial}; and recognizing if humans require assistance in schools, hospitals, or the workplace~\citep{ahamed2018applying,al2020survey,kulkarni2014healthcare}.

While the field of machine learning for IoT has great potential, existing efforts are often specialized to a single sensory modality or prediction task~\citep{atmoko2017IoT,khan2018real,kumar2020time,Cook2020anomaly}, resulting in limited resources to systematically study large-scale learning across many IoT sensors and tasks. To standardize the benchmarking and development of new machine learning technologies for IoT, this paper proposes \names, the most expansive and unified IoT benchmark to date, encompassing over $1.15$ million samples covering $12$ real-world sensory modalities and $8$ IoT tasks firmly rooted in practical scenarios such as personal wellness, healthcare, and smart cities. These tasks include perceiving the pose, gaze, activities, and gestures of humans as well as the touch, contact, pose, and 3D structure of physical objects. \names\ introduces the following unique challenges to the machine learning community:
\begin{enumerate}[noitemsep,topsep=0pt,nosep,leftmargin=*,parsep=0pt,partopsep=0pt]
    \item \textbf{High-modality multimodal learning}: While multimodal representation learning has historically been limited to image, text, video, and audio~\citep{liang2022foundations}, real-world sensory modalities like IMU, thermal dynamics, GPS, depth, camera captures, audio, and more paint a more realistic picture of our multisensory physical world. These diverse modalities introduce new challenges in generalization across modalities and multitask and transfer learning across different physical sensors.

    \item \textbf{Long-range temporal interactions}: The second challenge lies in learning fine-grained multimodal interactions across long temporal ranges. Real-world sensory data is naturally sequential, possibly over extremely long time ranges, and multisensory sequential data often shows interactions between time steps that are not aligned. For example, typical image-text datasets have a sequence length of 77 words or lower~\citep{lin2014coco}, video datasets are roughly 10-60 seconds long~\citep{zadeh2018multimodal}, while \names\ displays sequence lengths of up to 100-300 steps.

    \item \textbf{Heterogeneity and robustness}: The third challenge lies in handling the extreme heterogeneity in real-world sensors with unique structures and noise topologies~\citep{liang2021multibench,liang2022foundations}. These sensory modalities may be naturally noisy or corrupted, not easily semantically segmented, and may not have natural language correspondences like image and video often do.
    
    \item \textbf{Real-time complexity}: Finally, many IoT devices need to run in real-time for applications in smart cities, security, healthcare, and automation. We therefore need to benchmark the efficiency of multisensory data collection, processing, and prediction as a critical quality alongside performance.
\end{enumerate}

In addition to its diverse data modalities and prediction tasks, \names\ also contains a new set of evaluation metrics to study these challenges. Through this holistic benchmark, we evaluate a family of machine learning models spanning unimodal sensor-specific~\citep{Ahuja2021TouchPose,Riku2022RGBDGaze} and multisensor fusion approaches~\citep{lee2020multimodal,liang2022foundations,yeong2021sensor}, multimodal and multitask pre-training~\citep{jaegle2021perceiverio,liang2022highmmt,reed2022generalist}, and multimodal extensions of large language models~\citep{gao2023llamaadapterv2,zhu2023minigpt}.
Together, these cover all state-of-the-art frontiers of machine learning, deep learning, and foundation models for IoT. Our results highlight opportunities for ML to make a significant impact in IoT, but many challenges in scalable learning from heterogeneous, long-range, and imperfect sensory modalities are critical directions for future work.

Overall, \names\ presents a milestone in unifying disjoint efforts in machine learning and IoT research and paves the way towards a better understanding of the capabilities and limitations of current models, all the while ensuring ease of use, accessibility, and reproducibility. \names, evaluation metrics, standardized implementations of various models, and leaderboards are publicly available, will be regularly updated, and welcomes inputs from the community.

\vspace{-2mm}
\section{\names\ benchmark, modalities, and tasks}
\vspace{-2mm}

The rapid expansion of the Internet-of-Things (IoT) landscape necessitates a comprehensive benchmark that captures the richness and variety of IoT sensory modalities and tasks. \names\ is the largest and most diverse of its kind, comprising 1.15M samples spanning twelve distinct modalities and geared towards eight challenging tasks, as summarized in Figure~\ref{fig: title_img}.

\vspace{-2mm}
\subsection{Twelve diverse modalities}
\vspace{-2mm}

We collected diverse data from IoT devices, such as Inertial Measurement Units (IMU), Thermal sensors, Global Positioning Systems (GPS), capacitance, depth, gaze, and pose.
We also collect commonly used image, audio, and video modalities in the physical world to bridge conventional multimodal research with the new challenges introduced by \names.
\begin{enumerate}[noitemsep,topsep=0pt,nosep,leftmargin=*,parsep=0pt,partopsep=0pt]
    \item \textbf{Inertial measurement units} capture 3D motion and orientation. This data is fundamental for various applications, including motion tracking and navigation. We include 2,940 IMU gaze samples~\citep{kong2021EyeMU}, 28,400 IMU motion instances~\citep{Mollyn2022samosa}, 160,120 IMU samples~\citep{Riku2022RGBDGaze}, 330,178 IMU orientation recordings~\citep{Huang2018deep}, and 510,142 timestamps-based IMU samples~\citep{Grauman2022Ego4D}.

    \item \textbf{Thermal} provides temperature radiance insights, crucial in surveillance. We used 12,025 samples from LLVIP~\citep{jia2021llvip} containing pedestrians and cyclists from different locations on the street.

    \item \textbf{Global positioning systems} offer location data with high precision. This data is invaluable for tasks like location-based services, asset tracking, and navigation. We include GPS data from self-driving cars using 41,000 samples from KITTI~\citep{geiger2013kitti} using OXTS RT3003 inertial and GPS navigation system for depth estimation. The geographic coordinates include global orientation, altitude, velocities, accelerations, angular rates, and satellite information.

    \item \textbf{Cameras} capture the visual world in rich detail. We include 41,000 instances from KITTI self-driving car dataset~\citep{geiger2013kitti} using a Velodyne laser scanner installed on a vehicle car for depth estimation. The timestamp-based points can be considered according to the scanner's continuous rotation on its vertical axis, which provide context to GPS/IMU systems for auto-driving.

    \item \textbf{Capacitance} sensors measure changes in capacitance to detect nearby objects or changes and are critical components of touchscreen technologies and proximity sensing. We used 65,374 samples from TouchPose~\citep{Ahuja2021TouchPose} using a 39.6 cm capacitive Crystal Touch panel, 16-bit touch digitizer, and cameras. When fingers approach the lines on the mutual-capacitance touch sensor, it causes a capacitance drop between lines, resulting in the mutual-capacitance image.

    \item \textbf{Depth} sensors measure distances between the sensor and objects, providing a 3D view of the environment. They play a significant role in tasks like object detection and scene reconstruction. We used 160,120 samples from RGBGaze~\citep{Riku2022RGBDGaze} using Apple iPhone X with a TrueDepth camera.

    \item \textbf{Gaze} sensors track eye movement and direction, offering insights into user attention and intention. We used 2,940 samples from EyeMU~\citep{kong2021EyeMU} running an iOS application on an Apple iPhone 12 Pro. The participants were asked to gaze at a single red dot, and the screen advanced to capture a motion gesture and a 2-axis gaze location after 1.2 seconds.

    \item \textbf{Pose} sensors capture the orientation and position of objects or individuals critical for motion analysis and interactive applications. We include 330,178 samples from DIP-IMU~\citep{Huang2018deep} using Xsens IMU sensors, and 65,374 samples in TouchPose~\citep{Ahuja2021TouchPose} from a Leap Motion stereo IR camera, running Orion 4.1 for 3D hand pose tracking.

    \item \textbf{LiDAR} sensors emit light to measure distances, generating high-resolution 3D maps of environments. They are central to autonomous driving and topographical mapping. We include 51,000 samples from the Newer College dataset~\citep{Ramezani2020newer} using the Ouster LiDAR with 64 beams, 64 Channels, 120 m range, 45$^\circ$ vertical Field-of-View (FoV), and 1024 horizontal resolution.

    \item \textbf{Video} captures sequences of visual frames, providing a dynamic view of the environment. We used 510,142 egocentric videos in Ego4D~\citep{Grauman2022Ego4D}, which include many everyday activities, such as cooking, cleaning, and fishing from diverse geographic locations across the world, and are paired with timestamps-based IMU values of the normalized accelerometer and gyroscopes.

    \item \textbf{Audio} sensors capture sound waves, enabling voice recognition, sound classification, and environmental sound analysis. We include 28,400 samples from SAMoSA~\citep{Mollyn2022samosa}, where participants wore the smartwatch on their dominant arm, and were asked to perform 26 activities across 4 contexts with each activity repeated 3 times within each context.

    \item \textbf{Image} sensors offer static visual captures of the environment, serving as a basis for a myriad of vision tasks. We collected 160,120 samples from RGBDGaze~\citep{Riku2022RGBDGaze} paired with gaze, depth, and IMU for gaze tracking, 41,000 samples from KITTI~\citep{geiger2013kitti},  12,025 high-quality images paired with infrared thermal samples in LLVIP~\citep{jia2021llvip}, and 65,374 instances from TouchPose~\citep{Ahuja2021TouchPose}.

\end{enumerate}

\vspace{-2mm}
\subsection{Eight challenging tasks}
\vspace{-2mm}

Upon these 12 modalities, our benchmark includes tasks that reflect real-world IoT challenges.
\begin{enumerate}[noitemsep,topsep=0pt,nosep,leftmargin=*,parsep=0pt,partopsep=0pt]
    \item \textbf{Gaze estimation:} This task is pivotal for human-computer interaction, driver monitoring, and virtual reality. Given RGB images of faces, depth, and IMUs, our goal is to predict the location (X/Y) for tracking the gazes of the person. This regression task requires multisensory understanding of long-range interactions between RGB images and depth and heterogeneity in IMUs.

    \item \textbf{Depth estimation} involves predicting the distance between the camera and each pixel in the image and is a cornerstone for AR/VR applications, robotics, and object detection. Given RGB images, camera parameters, GPS coordinates, and IMU, we predict the depth maps of objects, such as cars and pedestrians on the streets. For robots, given RGB images, capacitive images, and hand poses, our target is to estimate the depth maps of left and right hands. 

    \item \textbf{Gesture classification:} Crucial for human-machine interfaces, gesture classification aims to recognize specific human hand or body movements. Given gaze locations and IMU data on accelerometer, gyroscope, and orientation, the goal is to classify human gestures. This classification problem requires the cross-modal perception of both gaze and IMUs.

    \item \textbf{Pose estimation} focuses on determining the spatial arrangement of human joints and has applications in AR/VR, gaming, and health. Given RGB images and measured IMU data, our goal is to predict the poses of the human body including 24 joints with three joint angles (yaw, pitch, roll). This regression problem requires fusing IMUs and RGB pixels.

    \item \textbf{Touch contact classification} involves determining the type or nature of touch on capacitive surfaces, a vital component for enhancing user experiences on touch-based devices. Given RGB images, capacitive images, depth maps, and hand poses, the goal is to classify touch contact.

    \item \textbf{Event detection:} A broad area with applications in health, wellness, smart homes, and the workplace, event detection involves identifying specific occurrences or anomalies in the data stream. Given audio spectrograms and IMU data of accelerometer, gyroscope, and orientation, our goal is to predict the categories of events across different timestamps. This classification problem requires modeling interactions between audio and IMU.

    \item \textbf{Activity recognition:} Central to fitness, health, and elderly care, activity recognition aims to discern human activities like walking, running, or jumping given RGB images, poses with three joint angles (yaw, pitch, roll), IMU data, or egocentric videos.

    \item \textbf{3D reconstruction} involves creating a three-dimensional model of an environment or object from 2D data, an application of huge significance in gaming, film, and AR/VR. Given RGB images, capacitance images, and depth maps, we aim to reconstruct 3D poses.

\end{enumerate}

\section{Modeling Paradigms in \names}

The models we include in \names\ span conventional IoT processing methods, which we briefly review below, as well as new methods we designed based on large multisensory foundation models.

\textbf{Domain-specific unimodal models.}
Over the years, each sensor modality has evolved its own set of algorithms. 
For instance, IMU data has been traditionally processed using Kalman filters~\citep{lai2019iot} to predict movement, while thermal modality often relies on image processing techniques for hotspot detection~\citep{George2018agraph,zhu2021hotspot,Chockalingam2023sensor}. 
A majority of IoT sensor data is inherently time-series~\citep{atmoko2017IoT,khan2018real,kumar2020time,Cook2020anomaly}.
Classical statistical methods like AutoRegressive Integrated Moving Average (ARIMA)~\citep{LOPEZMARTIN2020iot,Yasnita2020ahybrid} or Exponential Smoothing have been employed to forecast, denoise, or detect anomalies in sensor readings~\citep{Gardner1985exponential,Baki2006exponential,Alysha2011forecasting}. 
Signal processing methods, such as Fourier~\citep{zhang2017channel,Murugan2017an} and Wavelet Transforms~\citep{Muthukrishnan2019internet}, to data compression~\citep{Hossain2019iot} and feature extraction strategies specific to resource-constrained devices~\citep{Ebrahimi2019post,khor2021public,Imteaj2022a} have also been proposed.
Many of these methods were designed to function efficiently in real-time scenarios with limited computational resources.

\textbf{Multitask unimodal models} extend unimodal models by having a common backbone for the sensory modality and separate decoder heads, each suitable for predicting a single task~\citep{caruana1997multitask}. The common backbone can learn general-purpose information about the sensory modality while each decoder is task-specific. Given a dataset \( D = \{ (x_i, y_{i1}, y_{i2}, ... y_{in}) \} \) where each \( x_i \) has multiple corresponding labels for different tasks, the model minimizes a combined loss \( L \): 
\begin{equation}
L(D, M) = \sum_{i} \sum_{j} \mathcal{L}_j(M_j(E(x_i)), y_{ij}).
\vspace{-0.5em}
\end{equation}
where $M_j(\cdot)$ denotes the $j$th task model, and $E$ denotes the encoders.

\begin{figure}[t]
\centering
\vspace{-4mm}
\includegraphics[width=0.93\linewidth]{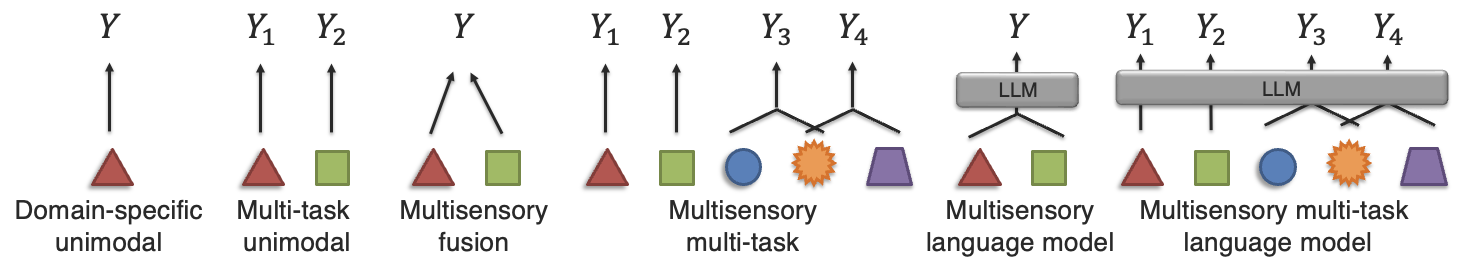}
\vspace{-2mm}
\caption{\names\ includes a suite of benchmark models spanning (1) domain-specific unimodal models using IoT expert knowledge, (2) multitask unimodal models with task sharing for each modality, (3) multisensory fusion models for single tasks, (4) multisensory multitask models that share information across many modalities and tasks, (5) multisensory language models that ground pretrained language models on sensor modalities, and (6) multisensory multitask language models grounded on sensor modalities for many tasks simultaneously.}
\vspace{-2mm}
\label{fig: main_img}
\end{figure}

\textbf{Multisensory fusion models} combine different modalities at some stage in the model – be it early fusion, middle fusion, or late fusion~\citep{liang2022foundations}. A common approach is to use separate encoders for each modality and a shared decoder that fuses the representations to produce an output.
Given multi-modal data \( x = (x_1, x_2, ... x_m) \), the model combines representations:
\begin{equation}
y = T(E_1(x_1) \oplus E_2(x_2) \oplus ... \oplus E_m(x_m))
\end{equation}
where $T(\cdot)$ denotes the task head, and $E_1, E_2, ..., E_m$ denote the encoders.

\textbf{Multisensory multitask models} leverage data from different modalities to solve more than one task simultaneously~\citep{jaegle2021perceiverio,liang2022highmmt,reed2022generalist}. It often benefits from interconnections between tasks. For example, in an IoT setting, a model could use vision and sound vision to simultaneously predict both the type of event occurring and its intensity, and further use vision and depth to reason about moving objects. For multi-modal data \( x = (x_1, x_2, ... x_m) \) and multiple tasks, the combined representations are processed as:
\begin{equation}
y_j = T_j(E_1(x_1) \oplus E_2(x_2) \oplus ... \oplus E_m(x_m))
\end{equation}
where $T_j(\cdot)$ denotes the $j$th task head, and $E_1, E_2, ..., E_m$ denote the encoders.

\textbf{Multisensory language models}: While the above approaches are primarily based on supervised learning across one or more modalities and tasks, there has been recent interest in grounding large language models on external modalities to take advantage of the general prediction, reasoning, and interaction capabilities of large language model decoders~\cite{touvron2023llama}. These methods operate via adapter layers that transform a modality's features into the original layers of a pre-trained model~\citep{gao2023llamaadapterv2}.
Given a pre-trained model with a set of weights \( W \), and an adapter module \( A \) with its own set of weights \( W_A \), the output \( y \) for an input \( x \) is:
\begin{equation}
y = M_{W+A}(x) = M_W(A_{W_A}(x)).
\end{equation}
where $M_{W+A}(\cdot), M_W(\cdot)$ denotes the model with both weights $W,A$ and weights $W$.

\textbf{Multisensory multitask language models} are multitask extensions of single-task adapters, where general representations for many tasks are transformed into the layers of a pre-trained model~\citep{gao2023llamaadapterv2}. For example, in an IoT setting with multi-modal data \( x = (x_1, x_2, ... x_m) \), we are given a pre-trained model with a set of weights \( W \), and an adapter module \( A \) with its own set of weights \( W_A \), the output \( y \) for an input \( x \) to formulate the Multisensory Multitask Adapter as:
\begin{equation}
y = M_{W+A}(E_1(x_1) \oplus E_2(x_2) \oplus ... \oplus E_m(x_m)) = M_W(A_{W_A}(E_1(x_1) \oplus E_2(x_2) \oplus ... \oplus E_m(x_m))).
\end{equation}
where $M_{W+A}(\cdot), M_W(\cdot)$ denotes the model with both weights $W,A$ and weights $W$, and $E_1, E_2, ..., E_m$ denote the encoders for multiple tasks with multisensory data.

\vspace{-2mm}
\section{Experiments}\label{sec:experiment}
\vspace{-2mm}

Our experiments aim to benchmark existing machine learning paradigms on \names, including the best task-specific models as well as those designed for multimodal, multitask, long-range, and noisy data settings. We elaborate on the experimental setup and report our findings.

\vspace{-1mm}
\subsection{Experimental Setup}
\vspace{-1mm}

All experiments were conducted on NVIDIA V100 GPUs. For unimodal models, data from each modality was processed independently using optimized neural architectures like CNNs for images and time-series models for sensor data. Models were trained with a batch size of 128, using the Adam optimizer at a learning rate of 0.001. Unimodal multitask models use shared encoder layers and task-specific decoders, and we ensured balanced gradients among tasks for equal training~\citep{liang2022highmmt}. For multisensory models, we experimented with specialized unimodal models with data fusion occurring at varying levels, from input to decision levels~\citep{liang2022foundations}. Multisensory multitask models utilized modality-specific encoders followed by task-specific decoders, enabling sharing across modalities and tasks during training. Finally, multisensory language models utilized adapter-based methods such as LLaMA-adpater~\citep{gao2023llamaadapterv2}, which enables us to keep the LLM frozen and only fine-tune only the small adapter modules, and multisensory multitask language models extend adapter-based fine-tuning to multiple tasks at the same time.

To evaluate performance, we employ task-specific metrics following prior practice.
For gaze and pose estimation, we measure the mean euclidean error in centimeters between predictions and ground truth. Depth estimation utilizes mean absolute error in millimeters, while gesture classification, touch contact classification, and activity recognition rely on accuracy metrics. Event detection employs the F1 score for confident threshold predictions, and 3D pose reconstruction is assessed using the End-point-error in millimeters for joint discrepancies.

\input{SECTIONS/40_Experiments/exp_sota}

\vspace{-2mm}
\subsection{Main quantitative results}
\vspace{-1mm}

\textbf{Overall performance}: Table~\ref{tab: exp_sota} reports the quantitative results on \names\ using single modality, single task, multimodal multitask, and extensions of language models.
As seen in Table~\ref{tab: exp_sota}, the multimodal multitask method consistently outperforms the single modality and single task models across all tasks.
This can be attributed to their ability to integrate information across modalities and tasks, which is especially crucial when one modality might have noisy or incomplete data.
While the multisensory language model often falls short in many scenarios, the multisensory multitask language model achieves the strongest results by leveraging the power of both multimodal inputs and multitask training, with the existing reasoning ability present in pretrained large language models.

\input{SECTIONS/40_Experiments/exp_modality}

\textbf{Performance across different modalities}: In this section, we study the impact of adding more modalities on task performance. Table~\ref{tab: exp_modality} shows significant performance improvements when adopting a multimodal approach as opposed to unimodal setups and various ratios (25\%, 50\%, all) of total modalities. 
The incorporation of multiple modalities results in more robust and accurate models. 
This can be attributed to the model's ability to tap into complementary information present in different modalities, especially in scenarios where one modality might be ambiguous or noisy.

\input{SECTIONS/40_Experiments/exp_task}

\input{SECTIONS/40_Experiments/exp_zeroshot}

\textbf{Performance across different tasks}: We separately analyze model performance when trained on multiple tasks simultaneously, while keeping the same modality inputs constant.
Table~\ref{tab: exp_task} reveals that for most tasks, our multitask model's performance was on par with or exceeded that of models trained solely on individual tasks. 
This suggests that the shared representations learned during multitask learning were largely beneficial,
since the model learns more generalized and robust features, while also improving computational efficiency.

\textbf{Zero-shot and few-shot transfer}: 
Furthermore, we study whether models trained on certain modalities or tasks can transfer to a new set of target modalities or tasks they have never seen during training (zero-shot) or have seen with only very few examples (few-shot 5, 10, 20). 
We chose the fix-8 dataset as the target, primarily because of its diverse representation of modalities (IMU, capacitance, depth, image) and its challenging task (gaze estimation and touch contact classification).
We examined various configurations ranging from transferring unimodal and multimodal multitask models.
From the results in Table~\ref{tab: exp_zeroshot}, we find that even zero-shot performance from a transferred multimodal multitask model can be comparable to supervised training using only IMU, depth, and image modalities. Furthermore, adding just a few examples (5-20) significantly boosted performance compared to the zero-shot setting, which highlights the model's ability to quickly learn new information.
Our results suggest that multimodal and multitask training enables few-shot capabilities that can be helpful for limited-data real-world IoT scenarios.

\vspace{-2mm}
\subsection{Understanding challenges in \names}
\vspace{-1mm}

\input{SECTIONS/40_Experiments/exp_longrange}

\textbf{Long-range multimodal interactions} are critical to many problems in IoT, such as in time series forecasting and signal analysis.
In a controlled experiment, we truncated sequences to various lengths and observed how conventional models performed. 
From Figure~\ref{fig: ablations} (left), as the sequence lengths increased, representing longer durations of time or more extensive contexts, there was a marked decline in performance. 
This showcased the models' inability to effectively encapsulate and understand interactions beyond a certain range.
Multimodal setups further complicate this when the long-range dependencies aren't just within a modality but can also be across modalities. 
Therefore, architectures that can handle both long-range and multisensory data will be critical for progress.

\textbf{Heterogeneity in structure}: Differences in data distributions, due to their natural structure, are a challenge for machine learning models.
We evaluated the same models on datasets that combined structured data (such as GPS, IMU) with unstructured data (such as images or raw audio) and found that unimodal baselines often struggled to reconcile these different data forms, leading to a significant drop in accuracy. 
The main finding from our evaluation is that when models are not tailored to the specific characteristics of each data type, their ability to effectively integrate and interpret data diminishes. The results in both gaze estimation and touch contact classification drop, underscoring the inadequacy of generic models in handling complex, mixed-data scenarios.
Therefore, the use of modality-specific encoders is critical in addressing the challenges posed by heterogeneous data. How to best handle these high degrees of heterogeneity, while maintaining efficiency beyond independent models for each sensor, is a critical direction for future work.

\textbf{Robustness to noisy and missing sensors}: 
In real-world applications, machine learning models often encounter data that is incomplete or corrupted by noise. 
To assess the robustness of our models against such imperfections, we introduced varying degrees of Gaussian noise into the datasets and systematically dropped sensor data at regular intervals. Both of these types of noise can naturally happen in real-world settings due to white noise and sensor failures respectively.
From Figure~\ref{fig: ablations} (right), we can observe the model's performance as we incrementally increase the noise ratio from 0\% to 50\%. At 0\% noise, the models operate in optimal conditions, showing peak accuracy. As the noise level increases to 10\% and 20\%, there is a noticeable degradation in performance, illustrating the initial sensitivity to noise. Beyond 20\%, the decline becomes more pronounced, with model accuracy dropping below 80\% at a 50\% noise ratio.
In addition to noise, missing sensor data is another common issue, and we find similar patterns when randomly omitting readings from various sensors. These findings indicate that building robust models for IoT is still a challenge.

\input{SECTIONS/50_Ablation_Study/ab_complexity}

\textbf{Complexity during training and inference}: 
One final critical considerations in the development of IoT models is the balance between the model's performance and its computational cost. 
Table~\ref{tab: ab_complexity} reports a comparative analysis of the performance across different methods against their respective training costs.
There is a clear incremental increase in performance from unimodal to multisensory multitask approaches. 
The unimodal method, while the least costly in terms of training time (25 hours), offers the lowest performance across all three tasks including touch classification, event detection, and activity recognition.
The shift towards multisensory multitask learning slightly increases the training costs but also yields notable enhancements in performance.
Overall, the multisensory multitask model yields the best tradeoffs between performance and complexity.

\input{SECTIONS/40_Experiments/vis_image}

\vspace{-2mm}
\subsection{Analysis of information sharing}
\vspace{-1mm}

Finally, we show visualization examples of how information is shared across modalities and tasks in Figure~\ref{fig: vis_img}, based on low-level modality features and high-level semantic concepts.

\textbf{Low-level modality features}:
Different sensory modalities often contain unique low-level perceptual features that complement those in other modalities. We illustrate this information sharing across 3 modalities: IMU, video, and pose data for predicting 2 common activities: walking and dancing.

\textit{Walking} is a common activity with distinctive rhythmic characteristics. Using IMU features, the model learns that rhythmic patterns, particularly in acceleration and deceleration, correspond to each walking step. The cadence, stability, and any irregularities in the walking pattern can also be inferred. Video features capture the holistic visual representation of walking, presenting details such as gait, arm swing, speed, stride length, and frequency. Finally, pose features highlight the specific posture changes during walking, emphasizing leg movement, foot placement, and body alignment.

\textit{Dancing} requires complex and expressive motions with varying styles and dynamics. IMU provides dynamic, often non-linear patterns in IMU data, reflecting the dance's tempo, vigor, and style variations; video captures the dance form, style, synchronization, and expressiveness; and pose data captures the alignment and configuration of body parts, offering insights into dance postures, transitions, and intricate footwork or hand movements.

\textbf{High-level semantic concepts}
encapsulate a more general conceptual understanding and reasoning about the environment. We show two examples showing how the audio and IMU modalities share information about two high-level semantic concepts, focusing on body pose and hand pose.

\textit{Body pose} represents the spatial arrangement and posture of the entire human body. This can involve stances like standing, sitting, or lying down, or even dynamic movements like jumping or running. For Audio, indirect cues such as the sound of footsteps, a person sitting down on a chair, or even the echo in a room (indicating a certain body pose affecting sound propagation) can provide hints about the body's posture.
For IMU, accelerometers capture the directional movement while gyroscopes provide rotational dynamics to distinguish if a person is upright, moving rapidly, or stationary.

\textit{Hand pose} looks at the orientation, gesture, and spatial arrangement of just the hands, ranging from gestures like waving and gripping, to more intricate signs in sign language. In audio, sounds like clapping, snapping, or even the subtle rustling of hands moving through the air can be detected. The distinct sounds made by hands interacting with other objects can also hint at specific hand poses. When IMU sensors are placed on the wrist or back of the hand, they can capture detailed dynamics of hand movements, tilting, rotation, or swift movements.

\vspace{-2mm}
\section{Conclusion and Broader Impacts}\label{sec:conclusion}
\vspace{-2mm}

This paper proposes \names, the most expansive IoT benchmark to date, encompassing over 1.15 million samples from 12 modalities and 8 tasks.
\names\ introduces unique challenges involving (1) learning from many sensory modalities, (2) fine-grained multisensory interactions across long temporal ranges, and (3) extreme heterogeneity due to ambiguous semantic abstraction and unique noise topologies in real-world sensors, which inspire several directions for future not encountered in conventional representation learning research.
\names, our standardized code, and leaderboards are publicly available, will be regularly updated, and welcome inputs from the community.

We are also aware of some potential \textbf{limitations and broader societal impacts}:
\begin{enumerate}[noitemsep,topsep=0pt,nosep,leftmargin=*,parsep=0pt,partopsep=0pt]
    \item \textbf{Data privacy}: There may be privacy risks associated with making predictions from multimodal data of recorded human behaviors, such as video, audio, activities, poses, and wearable sensors. Datasets are collected from participants who have consented to data release. We only use these datasets for research purposes. All data was anonymized and stripped of all personal (e.g., personally identifiable information) and protected attributes (e.g., race, gender).
    \item \textbf{Real-world privacy}: To deploy these algorithms at scale in the real world, it is also important to keep data and features private on each device without sending it to other locations using techniques such as federated learning~\citep{DBLP:journals/corr/abs-1812-06127,liang2020think}, differential privacy~\citep{geyer2017differentially}, or encryption~\citep{dankar2013practicing}. \names\ also enables large-scale studies of privacy-preserving machine learning in the IoT domain, which will be a critical direction for future work.
    \item \textbf{Efficiency}: Modern ML models can cause environmental impacts resulting from the carbon footprint required to run large-scale models. ML for IoT can inspire the design of lightweight models that can run efficiently on edge devices and low-cost sensors~\citep{strubell2019energy}.
    \item \textbf{Biases}: We also acknowledge risks of exposure bias due to imbalanced datasets, especially when human-centric data and possibly sensitive labels are involved. Models trained on biased data have been shown to amplify the underlying social biases~\citep{DBLP:journals/corr/abs-1809-07842}. Future work should quantify the internal learning process of multimodal models~\citep{liang2023multiviz} to better understand and mitigate social biases across sensory modalities. \names\ can be a useful resource to accelerate the study of fairer representation learning methods on real-world sensors.
\end{enumerate}

\newpage

{\footnotesize
\bibliography{reference}
\bibliographystyle{plainnat}
}

\newpage

\appendix

\section*{Appendix}

\input{SECTIONS/Appendix/manuscript}


\end{document}

%% file: math_commands.tex
\usepackage{amsmath,amsfonts,bm}

\def\eqref#1{equation~\ref{#1}}

\def\1{\bm{1}}

\DeclareMathAlphabet{\mathsfit}{\encodingdefault}{\sfdefault}{m}{sl}
\SetMathAlphabet{\mathsfit}{bold}{\encodingdefault}{\sfdefault}{bx}{n}



%% file: SECTIONS/10_Introduction/title_image.tex
\begin{figure}[!htb]
\centering
\vspace{-0mm}
\includegraphics[width=0.9\linewidth]{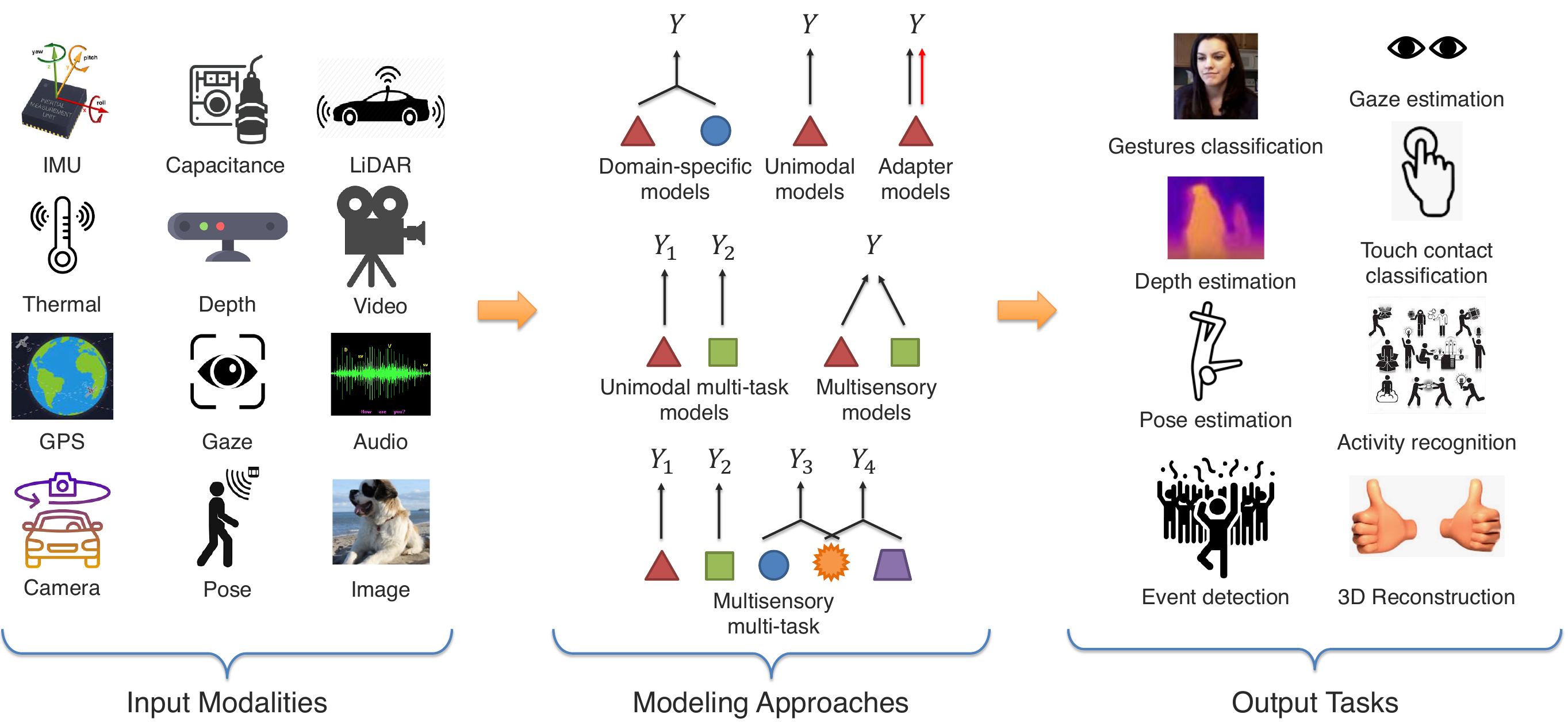}
\vspace{1mm}
 \caption{\names\ is the largest benchmark for machine learning on the Internet of Things (IoT), consisting of 1.15M samples, 12 rich modalities, and 8 challenging tasks such as perceiving the pose, gaze, activities, and gestures of humans as well as the touch, contact, pose, and 3D structure of physical objects. \names\ presents new challenges of (1) generalizable learning from many sensory modalities, (2) fine-grained interactions across long temporal ranges, (3) extreme heterogeneity and noise topologies in real-world sensors, and (4) complexity during training and inference.}
\vspace{-3mm}
\label{fig: title_img}
\end{figure}

%% file: SECTIONS/40_Experiments/exp_sota.tex
\begin{table}[t]
\centering
\caption{Multisensory multitask learning and multisensory multitask large language models are particularly effective approaches on \names, enabling information sharing to learn general representations for IoT data.}
\label{tab: exp_sota}
\scalebox{0.7}{
\begin{tabular}{lcccccccc}
    \toprule
    \multirow{2}{*}{Method} & Gaze est. & Depth est. & Gesture cls. & Pose est. & Touch cls. & Event det. & Activity recog. & 3D recons. \\
    & (cm, $\downarrow$) & (mm, $\downarrow$) & (\%,$\uparrow$) & (cm, $\downarrow$) & (\%, $\uparrow$) & (\%, $\uparrow$) & (\%, $\uparrow$) & (mm, $\downarrow$) \\
    \midrule
    Unimodal model & 2.26 & 20.7 & 97.3 & 6.49	  & 88.0 & 86.9 & 79.2 & 22.2 \\
    Unimodal multitask model & 1.95 & 18.2 & 98.2 & 5.36 & 89.3 & 88.1 & 82.5 & 20.5 \\
    Multisensory model & 1.79 & 17.3 & 98.7 & 4.62	  & 91.2 & 89.1 & 83.5 & 19.6 \\
    Multisensory multitask model & 1.08 & 13.6 & 99.3 & 3.85 & 93.8 & 92.7 & 87.5 & 17.5 \\
    Multisensory LM & 2.05 & 18.6 & 97.6 & 5.75	  & 88.7 & 87.5 & 82.3 & 21.3 \\
    Multisensory multi-task LM & \bf 0.95 & \bf 11.5 & \bf 99.6 & \bf 3.24 & \bf 94.6 & \bf 93.8 & \bf 89.2 & \bf 16.3 \\
    \bottomrule
\end{tabular}}
\vspace{-2mm}
\end{table}

%% file: SECTIONS/40_Experiments/exp_modality.tex
\begin{table}[t]
\centering
\caption{Adding more modalities enables complementary learning of information and yields improving performances on the \names\ benchmark.}
\label{tab: exp_modality}
\scalebox{0.75}{
\begin{tabular}{ccccccccc}
    \toprule
    \multirow{2}{*}{Modality Ratio} & Gaze est. & Depth est. & Gesture cls. & Pose est. & Touch cls. & Event det. & Activity recog. & 3D recons. \\
    & (cm, $\downarrow$) & (mm, $\downarrow$) & (\%,$\uparrow$) & (cm, $\downarrow$) & (\%, $\uparrow$) & (\%, $\uparrow$) & (\%, $\uparrow$) & (mm, $\downarrow$) \\
    \midrule
    single-modality  & 2.26 & 20.7 & 97.3 & 6.49 & 88.0 & 86.9 & 79.2 & 22.2 \\
    25\% & 2.13 & 19.6 & 97.5 & 5.97 & 88.9 & 87.3 & 80.2 & 21.5 \\
    50\% & 1.95 & 18.7 & 98.1 & 5.38 & 90.1 & 88.2 & 81.3 & 20.9 \\
    all & \bf 1.79 & \bf 17.3 & \bf 98.7 & \bf 4.62 & \bf 91.2 & \bf 89.1 & \bf 83.5 & \bf 19.6 \\
    \bottomrule
\end{tabular}}
\vspace{-3mm}
\end{table}

%% file: SECTIONS/40_Experiments/exp_task.tex
\begin{table}[t]
\centering
\caption{Multi-task learning is another effective strategy on the \names\ benchmark, enabling information sharing across tasks. Performance consistently improves as more datapoints from related tasks are added during training.}
\label{tab: exp_task}
\scalebox{0.75}{
\begin{tabular}{ccccccccc}
    \toprule
    \multirow{2}{*}{Task Ratio} & Gaze est. & Depth est. & Gesture cls. & Pose est. & Touch cls. & Event det. & Activity recog. & 3D recons. \\
    & (cm, $\downarrow$) & (mm, $\downarrow$) & (\%,$\uparrow$) & (cm, $\downarrow$) & (\%, $\uparrow$) & (\%, $\uparrow$) & (\%, $\uparrow$) & (mm, $\downarrow$) \\
    \midrule
    single-task  & 2.26 & 20.7 & 97.3 & 6.49 & 88.0 & 86.9 & 79.2 & 22.2 \\
    25\% & 2.17 & 19.9 & 97.5 & 6.23 & 88.3 & 87.1 & 80.1 & 21.8 \\
    50\% & 2.09 & 19.0 & 97.8 & 5.86 & 88.9 & 87.5 & 81.2 & 21.2 \\
    all & \bf 1.95 & \bf 18.2 & \bf 98.2 & \bf 5.36 & \bf 89.3 & \bf 88.1 & \bf 82.5 & \bf 20.5 \\
    \bottomrule
\end{tabular}}
\end{table}

%% file: SECTIONS/40_Experiments/exp_zeroshot.tex
\begin{table}[t]
\centering
\caption{Multimodal and multitask training enables zero-shot and few-shot capabilities, which can help when dealing with limited labeled data often seen in real-world IoT systems.}
\label{tab: exp_zeroshot}
\scalebox{0.75}{
\begin{tabular}{lcc}
    \toprule
    Method & Gaze estimation (cm, $\downarrow$) & Touch contact classification (\%, $\uparrow$) \\
    \midrule
    IMU & 2.65 & -- \\
    capacitance & -- & 83.5 \\
    depth & 2.45 & 86.2 \\
    image  & 2.26 & 88.0 \\
    multimodal & 1.79 & 91.2 \\
    multimodal multitask & \bf 1.08 & \bf 93.8 \\ \hline
    multimodal multitask (zero-shot) & 2.18 & 88.6 \\
    multimodal multitask (5-shot) & 1.96 & 89.5 \\
    multimodal multitask (10-shot) & 1.89 & 90.2 \\
    multimodal multitask (20-shot) & \bf 1.81 & \bf 91.1 \\
    \bottomrule
\end{tabular}}
\vspace{-4mm}
\end{table}

%% file: SECTIONS/40_Experiments/exp_longrange.tex
\begin{wrapfigure}{r}{0.6\textwidth}
\centering
\vspace{-4mm}
\includegraphics[width=0.92\linewidth]{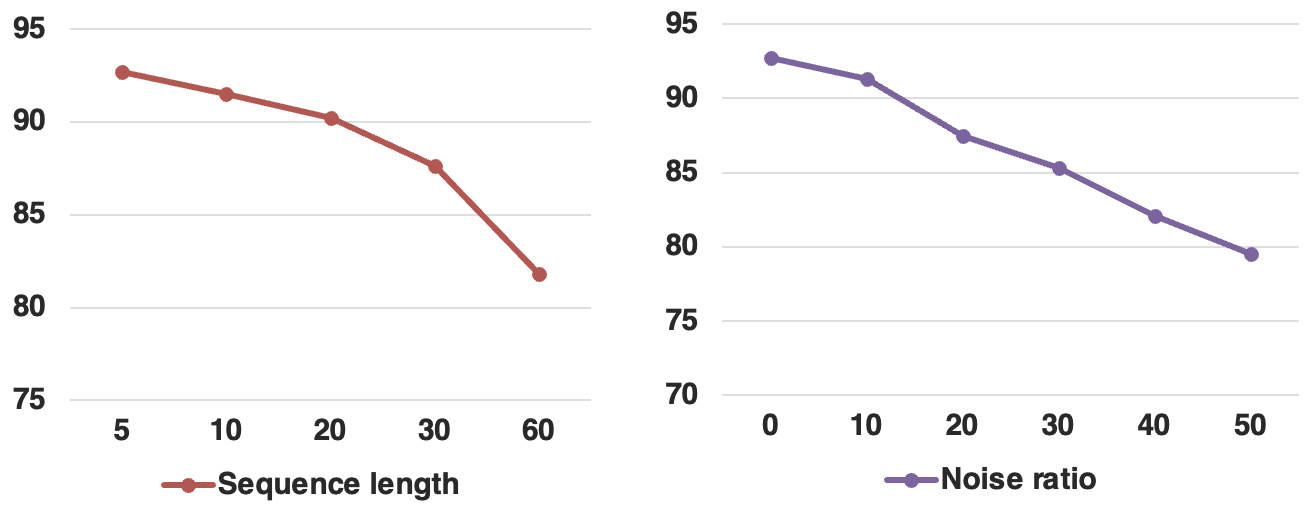}
\vspace{-0mm}
\caption{Long-range multimodal interactions and heterogeneity between modalities due to noise and imperfections make the \names\ benchmark particularly challenging for machine learning models.}
\vspace{-0mm}
\label{fig: ablations}
\end{wrapfigure}

%% file: SECTIONS/50_Ablation_Study/ab_complexity.tex
\begin{table}[t]
\centering
\vspace{-4mm}
\caption{Tradeoff between various models in terms of performance and training cost on \names. Multisensory multi-task models yield stronger performance but come at the expense of increased training costs.}
\label{tab: ab_complexity}
\scalebox{0.75}{
\begin{tabular}{lcccc}
    \toprule
    \multirow{2}{*}{Method} & Touch cls. & Event det. & Activity recog. & Average training cost  \\
    & (\%, $\uparrow$) & (\%, $\uparrow$) & (\%, $\uparrow$) & (hours,$\downarrow$) \\
    \midrule
    Unimodal model & 88.0 & 86.9 & 79.2 & \bf 25 \\
    Unimodal multi-task model & 89.3 & 88.1 & 82.5 & 30 \\
    Multisensory model & 91.2 & 89.1 & 83.5 & 32 \\
    Multisensory multi-task model & 93.8 & 92.7 & 87.5 & 38\\
    Multisensory language model & 88.7 & 87.5 & 82.3 & 27 \\
    Multisensory multi-task language model & \bf 94.6 & \bf 93.8 & \bf 89.2 & 39 \\
    \bottomrule
\end{tabular}}
\vspace{-4mm}
\end{table}

%% file: SECTIONS/40_Experiments/vis_image.tex
\begin{figure}[t]
\centering
\includegraphics[width=0.9\linewidth]{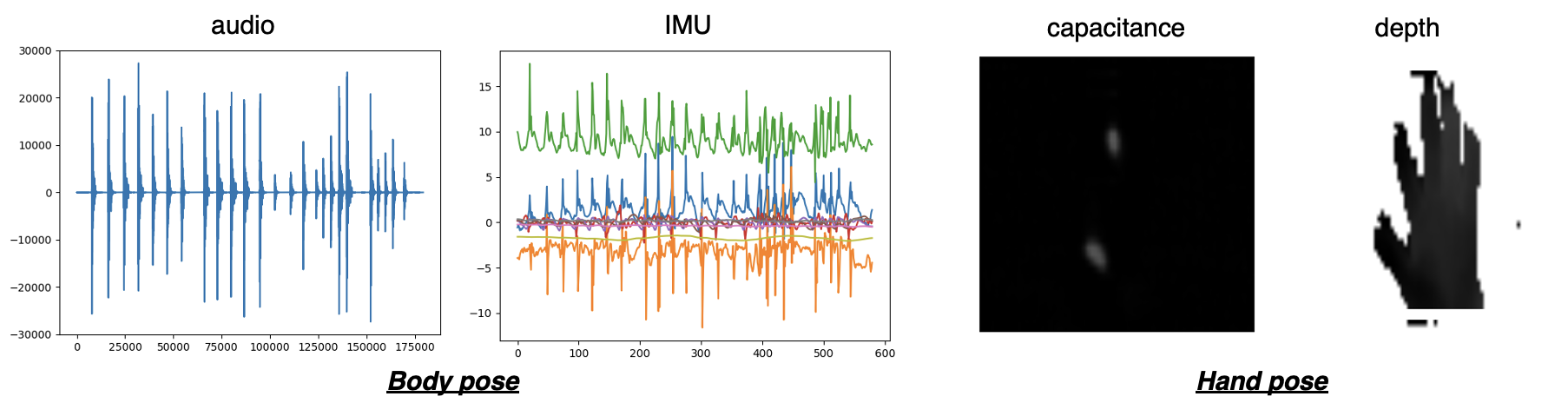}
\vspace{-2mm}
\caption{Visualizations of information sharing across body pose and hand pose on low-level modality features and high-level semantic concepts regarding audio, IMU, capacitance, and depth.
The audio and IMU modalities share the same concept of walking in body pose, while the capacitance and depth modalities share the concept of gripping in hand pose.
}
\label{fig: vis_img}
\vspace{-4mm}
\end{figure}

%% file: SECTIONS/Appendix/manuscript.tex
\appendix

In this supplementary material, we provide the following material:
\begin{itemize}
    \item addition implementation and datasets details in Section~\ref{sec: data_appendix},
    \item detailed experimental setup in Section~\ref{sec: imple_appendix},
    \item details about evaluation metrics in Section~\ref{sec: metric_appendix},
    \item more experimental analyses in Section~\ref{sec: exp_appendix},
    \item more qualitative visualization results in Section~\ref{sec: vis_appendix},
    \item more dialog examples for language models in Section~\ref{sec: lm_appendix},
    \item dataset documentation and intended uses in Section~\ref{sec: data_doc}.
\end{itemize}

\section{Detailed Benchmark}\label{sec: data_appendix}

We introduce the \names\ benchmark, the largest and most diverse IoT dataset consisting of 1.15 million samples across twelve distinct modalities, tailored towards eight challenging tasks.

\vspace{-2mm}
\subsection{Technical challenges and selection criterion}
\vspace{-1mm}

In this section, we outline unique challenges and potential real-world applications of representation learning for IoT, highlighting how these differentiate from traditional approaches. Our selection of modalities and tasks is driven by these challenges, detailed subsequently with an in-depth description of the benchmark composition.

\begin{enumerate}[noitemsep,topsep=0pt,nosep,leftmargin=*,parsep=0pt,partopsep=0pt]
    \item \textbf{High Modality Complexity:} The primary challenge in IoT representation learning involves handling high-modality data from diverse sensors and environments. Unlike conventional multimodal research limited to visuals and audio, \names\ incorporates advanced modalities such as IMU sensors~\citep{Huang2018deep}, thermal dynamics~\citep{jia2021llvip}, GPS signals, and camera feeds. This variety ensures comprehensive simulation of the real-world, enhancing the robustness and applicability of the learned representations. The integration of these varied modalities necessitates innovative approaches in data fusion, representation learning, and generalization across heterogeneous sensor data.

    \item \textbf{Temporal Dynamics:} IoT devices often capture data that embodies complex temporal dynamics over extended periods. Unlike typical multimodal datasets with short sequence lengths (e.g., image-text datasets averaging 77 words~\citep{lin2014coco}, or video datasets spanning 10-60 seconds), \names\ includes data sequences up to 100-300 steps, representing a significant leap in capturing long-range temporal interactions. This aspect introduces challenges in modeling sequential interactions that are not temporally aligned, thereby pushing the boundaries of current sequence learning methods.

    \item \textbf{Real-world Variability:} The heterogeneity and inherent noise in IoT sensor data pose substantial challenges. \names\ encompasses sensors with diverse noise signatures and lacks straightforward natural language equivalents, complicating the direct application of conventional conditioning techniques used in language models. This aspect of the benchmark tests models' abilities to handle real-world data variability and encourages the development of techniques that enhance noise robustness and semantic interpretation.

    \item \textbf{Real-time Processing:} The real-time nature of IoT applications demands models that can process and react to multimodal inputs promptly. This requirement is crucial in fields like healthcare monitoring, home automation, and security systems. The benchmark, therefore, not only measures model accuracy and robustness but also emphasizes efficiency and speed, ensuring that the models are practical for real-time applications.

\end{enumerate}
Reflecting these four core challenges, \names\ aggregates data from a wide array of environments and IoT devices, offering an unparalleled resource for advancing IoT research. This benchmark sets new standards in the field by providing a robust platform for developing next-generation IoT technologies that can efficiently handle complex, real-time, and multi-modal data streams.

\vspace{-2mm}
\subsection{Twelve Rich Modalities}
\vspace{-1mm}

The \names\ Benchmark integrates a diverse array of modalities, spanning structured and unstructured data, each bringing unique technical challenges and offering distinct perspectives on sensor-based machine learning.

\textbf{IMU (Inertial Measurement Units):} IMUs provide 3D motion and orientation data crucial for applications such as motion tracking and navigation. The challenge lies in accurately interpreting the noisy signals from accelerometers and gyroscopes, which are often affected by drift and require sophisticated filtering techniques to derive precise readings. We incorporated a rich dataset of IMU samples from various sources including EyeMU~\citep{kong2021EyeMU} for gaze estimation and SAMoSA~\citep{Mollyn2022samosa} for synchronized 9-axis data, enhancing the benchmark's depth in capturing real-world motion.

\textbf{Thermal:} Thermal sensors capture temperature variations, essential for applications like surveillance. The primary challenge is processing the subtle thermal changes in diverse environmental conditions without being overwhelmed by background noise. Our collection includes 12,025 thermal images from LLVIP~\citep{jia2021llvip}, providing a basis for advanced thermal pattern recognition tasks.

\textbf{GPS:} This modality is critical for location-based services and navigation, where the challenge is to deal with signal occlusion and multipath propagation in urban settings. The benchmark includes 41,000 GPS samples from KITTI~\citep{geiger2013kitti}, offering a framework to develop and test algorithms that can robustly estimate location even in less-than-ideal conditions.

\textbf{Camera:} As a cornerstone of computer vision, cameras provide rich visual data but must contend with challenges such as varying lighting conditions, occlusions, and dynamic environments. Our dataset incorporates comprehensive camera data from KITTI, which is instrumental in tasks like depth estimation and object recognition.

\textbf{Capacitance:} These sensors detect touch and proximity by measuring changes in capacitance. One challenge is distinguishing between intentional touch and incidental contact, critical for touch-sensitive applications. We used data from TouchPose~\citep{Ahuja2021TouchPose}, which includes detailed interactions captured via a capacitive touch panel.

\textbf{Depth:} Depth sensors, which provide spatial data about the surroundings, are crucial for 3D modeling and interaction systems. The challenge is to derive accurate depth information in cluttered scenes or where depth cues are minimal. Our dataset includes depth data from RGBGaze~\citep{Riku2022RGBDGaze} and TouchPose, enhancing tasks related to 3D reconstruction and interaction.

\textbf{Gaze:} Tracking where a user is looking offers insights into user intent and focus. The variability in individual gaze patterns and external lighting conditions makes this data challenging to interpret. The benchmark features detailed gaze data collected using iOS devices, facilitating the development of personalized gaze-tracking technologies.

\textbf{Pose:} Pose data is essential for understanding body movements and interactions. Capturing accurate pose information involves challenges related to body occlusions and the 3D nature of human movements. We include extensive pose data from DIP-IMU~\citep{Huang2018deep} and TouchPose, providing a foundation for advanced pose estimation algorithms.

\textbf{LiDAR:} Used for generating precise 3D maps, LiDAR data is pivotal for autonomous vehicles and geographic mapping. The challenge is processing the massive point clouds efficiently, especially in dynamic environments. Our inclusion of LiDAR data from the Newer College dataset~\cite{Ramezani2020newer} enriches the benchmark's utility for high-resolution spatial analysis.

\textbf{Video:} Video data captures dynamic scenes and is crucial for understanding temporal variations. The challenge lies in processing high-volume data streams in real-time, crucial for applications like surveillance and live activity recognition. The Ego4D~\cite{Grauman2022Ego4D} dataset contributes extensive egocentric video data, pushing forward research in first-person visual understanding.

\textbf{Audio:} This modality is essential for speech and environmental sound analysis, with challenges in noise filtering and sound source separation. We integrate audio data from SAMoSA~\cite{Mollyn2022samosa}, which, paired with IMU data, enhances the capability to analyze sounds in context.

\textbf{Image:} Static images are fundamental for numerous vision tasks, and challenges include dealing with diverse image qualities and contexts. Our dataset includes high-quality images from various sources, supporting a wide range of image processing and analysis tasks.

These modalities, each with its inherent challenges, make the \names\ Benchmark a comprehensive toolkit for advancing IoT device capabilities and multimodal learning research.

\subsection{Eight Well-defined and Challenging Tasks}
\vspace{-1mm}

Our benchmark outlines tasks that encapsulate real-world IoT challenges, designed to propel advancements in practical applications with significant societal impacts.

\textbf{Gaze Estimation:}
Integral to enhancing human-computer interaction, driver monitoring, and immersive experiences in virtual reality, gaze estimation involves predicting the (X, Y) coordinates of a person's gaze based on multimodal inputs. Given RGB images of faces alongside depth information and IMU data, this regression task tests the model's ability to integrate and interpret visual cues with spatial dynamics, addressing the challenge of sensor heterogeneity.

\textbf{Depth Estimation:}
Critical for augmented reality, robotics, and autonomous driving, depth estimation requires predicting the distance from the camera to each image pixel. Utilizing inputs such as RGB images combined with camera parameters, GPS, and IMU data, models must generate precise depth maps. This task emphasizes understanding complex spatial relationships and integrating diverse sensory data, crucial for navigating and interacting with real-world environments.

\textbf{Gesture Classification:}
Essential for developing intuitive human-machine interfaces, this task involves recognizing specific human gestures using gaze data and IMU outputs. Models must classify movements effectively by synthesizing data from accelerometers, gyroscopes, and orientation sensors, showcasing the need for robust cross-modal integration.

\textbf{Pose Estimation:}
This task aims at determining the spatial configuration of human joints, crucial in gaming, AR/VR, and health monitoring. Given RGB images and IMU data, the challenge is to predict human poses, including detailed joint angles. The task demands deep cross-modal insights, especially in blending visual information with physical sensor data.

\textbf{Touch Contact Classification:}
In touch-based user interfaces, accurately identifying the nature of touch interactions on capacitive surfaces is vital. This classification task leverages RGB and capacitive images, depth maps, and hand poses to discern touch types, highlighting the importance of multimodal interactions and the complexity of synchronizing disparate data types.

\textbf{Event Detection:}
Widely applicable in surveillance and smart environments, this task requires the detection of specific events or anomalies from audio-visual streams. Using audio spectrograms and IMU data, models must discern and categorize events, a process that hinges on the model's ability to correlate audio signals with physical sensor outputs across varied temporal spans.

\textbf{Activity Recognition:}
Critical for applications in fitness and healthcare, this task involves recognizing human activities from multimodal data. Models are challenged to integrate RGB images, video frames, poses, and IMU data to classify actions accurately, demanding a nuanced understanding of both motion and visual cues in dynamic, real-time scenarios.

\textbf{3D Reconstruction:}
Important in entertainment and spatial computing, 3D reconstruction involves creating detailed 3D models from 2D data. Given RGB images, capacitance data, and depth maps, the task tests the model's ability to construct accurate 3D representations, requiring a sophisticated blend of image processing and depth perception skills.

Each task is formulated to push the boundaries of what's possible with current technology, encouraging innovation in the handling of complex, multimodal datasets. These tasks not only reflect pressing real-world challenges but also serve as a robust platform for developing next-generation machine learning models tailored for the IoT ecosystem.

\section{Experimental Setup}\label{sec: imple_appendix}

In this section, we provide details about the experimental configurations used to evaluate our models across various tasks and modalities, ensuring reproducibility and evaluation.

\subsection{Setup for Domain-specific Unimodal Models}
\begin{itemize}
    \item \textbf{Data Preparation:} Each modality (e.g., RGB images, capacitive images, hand pose) is independently processed. Normalization and modality-specific transformations are applied to standardize the input data for optimal model performance.
    \item \textbf{Network Architecture:} Tailored neural architectures are employed for each modality type. For instance, convolutional neural networks (CNNs) are used for image data and recurrent neural networks (RNNs) for sequential data, optimizing each model's capacity to extract relevant features effectively.
    \item \textbf{Training Details:} Models are trained with a batch size of 128 using the Adam optimizer at a learning rate of 0.001. Early stopping is implemented with a patience of 10 epochs to prevent overfitting.
    \item \textbf{Evaluation:} Performance is methodically evaluated on validation datasets specific to each modality, enabling direct assessment of model efficacy and generalization.
\end{itemize}

\subsection{Setup for Multi-task Unimodal Models}
\begin{itemize}
    \item \textbf{Data Preparation:} Data pertinent to different tasks, but from the same modality, are concatenated or paired, enriching the training set.
    \item \textbf{Network Architecture:} A shared encoder processes the unified input data, followed by multiple task-specific decoders tailored to address the requirements of each task independently.
    \item \textbf{Training Details:} Gradient balancing techniques are utilized to ensure no single task dominates the learning process. This balanced approach is critical for maintaining uniform model performance across tasks.
    \item \textbf{Evaluation:} Each task is separately evaluated on tailored validation sets, highlighting the model’s task-specific competencies and areas for improvement.
\end{itemize}

\subsection{Setup for Multisensory Fusion Models}
\begin{itemize}
    \item \textbf{Data Preparation:} Different modalities are fused at various levels-input, feature, or decision—based on the nature of the task and the characteristics of the data.
    \item \textbf{Network Architecture:} Encoders specific to each modality process inputs independently before fusion layers integrate the features, aiming to capture and utilize the comprehensive information available across the modalities.
    \item \textbf{Training Details:} A uniform training approach using a batch size of 128 and the Adam optimizer is maintained, with particular attention to data balancing to ensure equitable representation from all modalities.
    \item \textbf{Evaluation:} The efficacy of the combined model is validated on a mixed-modality dataset, testing the model’s ability to synthesize and leverage multimodal information.
\end{itemize}

\subsection{Setup for Multisensory Multitask Models}
\begin{itemize}
    \item \textbf{Data Preparation:} A combination of data from various modalities and tasks is either paired or concatenated, depending on the specific requirements of each task.
    \item \textbf{Network Architecture:} Shared modality-specific encoders are followed by task-specific decoders, allowing fine-tuned processing paths for each task while leveraging shared learning across modalities.
    \item \textbf{Training Details:} The training regimen employs gradient and modality balancing techniques to foster a fair learning environment, promoting equal learning opportunities for all tasks and modalities.
    \item \textbf{Evaluation:} Task and modality-specific performance metrics are used to assess each aspect of the model’s capabilities comprehensively.
\end{itemize}

\subsection{Setup for Multisensory Language Models}
\begin{itemize}
    \item \textbf{Data Preparation:} All data is pre-processed to fit the input requirements of a pre-trained network, ensuring only the adapter modules are adaptable.
    \item \textbf{Network Architecture:} State-of-the-art architectures such as LLaMA are enhanced with adapter layers, strategically inserted to refine the model's ability to integrate and process multisensory data.
    \item \textbf{Training Details:} Adapter layers are fine-tuned with a larger batch size of 256, using the Adam optimizer at a reduced learning rate of 0.0005, optimizing for efficient learning dynamics.
    \item \textbf{Evaluation:} The performance of the adapted model is critically assessed against a validation set designed to challenge its enhanced capabilities, ensuring rigorous testing of its applied enhancements.
\end{itemize}

\subsection{Setup for Multisensory Multitask Language Models}
\begin{itemize}
    \item \textbf{Data Preparation:} Combines multisensory data streams with language data to prepare for multitask processing.
    \item \textbf{Network Architecture:} Integrates language processing units with sensory data processors within a unified architectural framework, facilitating complex multitask learning.
    \item \textbf{Training Details:} Employs sophisticated multitask learning algorithms to optimize performance across varied sensory and language tasks.
    \item \textbf{Evaluation:} Each task is individually assessed to determine the model's effectiveness across the spectrum of included tasks and modalities.
\end{itemize}

\noindent Throughout all experimental setups, the environment remains consistent. All models are trained and evaluated on NVIDIA V100 \& A100 GPUs, ensuring uniformity in computational power and performance, crucial for fair and replicable validation.

\section{Evaluation Metrics}\label{sec: metric_appendix}

To ensure a comprehensive assessment of model performance across diverse tasks, we employ a variety of metrics that are well-suited to the specific challenges posed by each task in our benchmark. These metrics not only align with industry standards but also facilitate a nuanced analysis of model effectiveness in real-world scenarios.

\textbf{Gaze Estimation:}
The precision of gaze tracking is quantified using the \textit{Mean Euclidean Error} in centimeters. This metric measures the average distance between the coordinates of the predicted gaze and the actual gaze points, providing a direct assessment of accuracy in spatial terms.

\textbf{Depth Estimation:}
For evaluating the accuracy of depth predictions, we use the \textit{Mean Absolute Error} (MAE) in millimeters. MAE helps quantify the average magnitude of errors in the predictions without considering their direction, making it particularly useful for depth where exact value prediction is crucial.

\textbf{Gesture Classification:}
The effectiveness of gesture recognition is measured by \textit{Accuracy}, defined as the ratio of correctly classified samples to the total samples. This metric is straightforward and reflects the model's ability to correctly identify and categorize each gesture, crucial for interactive applications.

\textbf{Pose Estimation:}
Similar to gaze estimation, pose accuracy is evaluated using \textit{Mean Euclidean Error} in centimeters. This metric calculates the average Euclidean distance between predicted and true joint positions, offering a clear measure of spatial accuracy in pose estimation.

\textbf{Touch Contact Classification:}
For assessing how accurately the model classifies types of touch interactions, \textit{Accuracy} is again utilized. This metric is particularly important for applications where precise touch recognition can enhance user interface responsiveness and interactivity.

\textbf{Event Detection:}
The \textit{F1 Score} is used to evaluate event detection, combining the measures of precision and recall. F1 is especially suitable for scenarios where a balance between false positives and false negatives is crucial, such as in surveillance or safety monitoring systems.

\textbf{Activity Recognition:}
We compute \textit{Balanced Accuracy} to evaluate activity recognition models. This metric is important in scenarios with imbalanced datasets, as it considers the accuracy of each class, thereby ensuring fairness across less frequent activities.

\textbf{3D Pose Reconstruction:}
The \textit{End-point Error} in millimeters assesses the precision of 3D pose reconstruction by measuring the mean Euclidean distance between all corresponding joints in the predicted and actual models. This metric is critical for applications in AR/VR and animation, where spatial accuracy in three dimensions is paramount.

Each of these metrics has been chosen to reflect both the efficacy and the practical utility of the models in handling real-world tasks, ensuring that our evaluations are both rigorous and relevant to practical applications.

\section{More Analysis}\label{sec: exp_appendix}

In this section, we delve deeper into two critical aspects of machine learning challenges that our experiments focused on: handling long-range interactions and managing heterogeneity in structure and noise.

\subsection{Testing Long-range Interactions}
Long-range interactions are essential for many machine learning applications, including time-series forecasting, natural language processing, and signal analysis. Recognizing patterns and relationships over extended sequences or across multiple modalities requires models that effectively leverage these long-range dependencies.
We conducted experiments in which we systematically truncated sequences to various lengths and analyzed the performance of conventional models. As sequence lengths increased, indicating longer durations or more extensive contexts, model performance declined noticeably, highlighting a deficiency in capturing and understanding distant interactions.

Complexity is amplified in multimodal contexts, where long-range dependencies exist not only within individual modalities but also across different modalities. Our findings indicate this area of intermodality long-range interaction is particularly challenging, with even advanced models showing limitations.
Developing architectures that inherently focus on long-range interactions, such as leveraging modified self-attention mechanisms capable of handling extremely long sequences.
Implementing models that operate at different temporal scales to summarize information effectively across various levels could enhance the capture of longer-range interactions.
Employing dynamic computational resource allocation techniques to emphasize critical parts of a sequence or modality when potential long-range dependencies are detected.
For multimodal problems, enhancing cross-modal attention mechanisms to enable models to recognize and utilize dependencies spanning across different modalities and temporal gaps.

\subsection{Testing Heterogeneity in Structure and Noise}
Heterogeneity in data, in terms of both structure and noise, poses significant challenges, especially as datasets become more complex and diverse.
Models were exposed to datasets combining structured data (such as GPS and IMU) with unstructured data (like images and raw audio). Unimodal baselines struggled significantly with these mixed data types, leading to notable accuracy reductions. Additionally, introducing varying degrees of Gaussian noise into numerical data demonstrated a rapid performance decline as noise levels increased.

These challenges underscore the need for enhanced robustness in model design. Our experiments reveal that even state-of-the-art models are vulnerable when confronted with unexpected data structures or noise patterns.
Exploring architectures and training strategies that inherently boost robustness to noise and heterogeneity, such as noise injection during training or using dropout techniques to foster generalization.
Applying advanced data augmentation techniques for both structured and unstructured data to better prepare models for diverse data structures and noise scenarios.
Utilizing meta-learning approaches to train models that can quickly adapt to new data structures or noise patterns with minimal additional training.
Developing sophisticated denoising mechanisms, including pre-processing methods and in-model techniques, particularly those capable of handling structured noise.
The detailed analysis and proposed solutions aim to direct future research toward developing machine learning models that are not only effective but also robust and adaptable to real-world data complexities.

\section{More examples}\label{sec: vis_appendix}

In this section, we provide more examples for an in-depth look at the diverse IoT modalities presented in our benchmark, emphasizing the heterogeneity in sensor types, data characteristics, and their implications for temporal interactions. These examples are instrumental in demonstrating the complex data processing challenges and the necessity for advanced interpretation techniques within IoT systems.

\input{SECTIONS/Appendix/vis_imu}

\subsection{IMU}

The Inertial Measurement Unit (IMU) is crucial for capturing dynamic motion and orientation, combining accelerometers, gyroscopes, and magnetometers. 
This modality is pivotal in devices like smartwatches, capturing high-resolution temporal data on user movement which is essential for applications in activity recognition and health monitoring. 
As shown in Figure~\ref{fig: vis_imu}, the IMU's high sampling rate allows for capturing minute fluctuations in motion, providing a detailed temporal analysis that is vital for accurate modeling of dynamic behaviors.

\input{SECTIONS/Appendix/vis_audio}

\subsection{Audio}

Audio sensors transform sound waves into digital signals, reflecting the acoustic environment. 
These sensors are extensively used in smart homes to detect various sounds—from spoken commands to household activity noises, as illustrated in Figure~\ref{fig: vis_audio}.
The detailed temporal granularity of audio data, essential for precise speech recognition, environmental sound classification, and emergency alert systems, showcases the critical role of temporal resolution in understanding and responding to audio cues effectively.

\input{SECTIONS/Appendix/vis_cap}

\subsection{Capacitance}

Capacitance sensing involves detecting changes in capacitance caused by touch or proximity. This technology enables the development of interactive touch interfaces and non-contact object and human interaction monitoring in IoT applications. 
The visualizations of capacitive data in Figure~\ref{fig: vis_cap} illustrate its responsiveness and the importance of real-time processing, crucial for applications requiring immediate interaction feedback.

\input{SECTIONS/Appendix/vis_depth}

\subsection{Depth}

Depth sensors measure distances using technologies like LIDAR, structured light, or time-of-flight cameras, critical for understanding spatial relationships in environments necessary for autonomous navigation or interactive applications. 
The visualizations in Figure~\ref{fig: vis_depth} highlight how depth data captures scene dynamics and object relations over time, underpinning the significance of temporal interactions in depth sensing for robust environmental understanding and interaction.

\input{SECTIONS/Appendix/vis_dialog}

\section{More Dialog Examples}\label{sec: lm_appendix}

In this section, we illustrate the analytical capabilities of our multisensory multi-task language model when applied to two distinct IoT modalities: audio and Inertial Measurement Unit (IMU) data. Through detailed dialog examples, we demonstrate the model's adeptness at interpreting heterogeneous sensory data and making informed inferences about activities based on these modalities.

\subsection{Audio Analysis}

\textbf{Context}:
In the audio analysis scenario, our model was presented with a visual representation of an audio waveform. The task was to identify the corresponding activity from a predefined list of categories based on the visualized audio pattern.

\textbf{Input}:
An audio waveform is visualized, showing varying amplitudes and frequencies over time.

\textbf{Task}:
The model is instructed to determine the activity that best matches the waveform from options such as "Talking," "Clapping," "Coughing," and "Music."

\textbf{Model Output}:
The multisensory multi-task language model identifies "Coughing" as the most likely activity.

\textbf{Rationale}:
The model justifies its choice by highlighting the sudden, sporadic spikes in the waveform, characteristic of the sharp, burst-like sounds typical of coughing. This decision demonstrates the model’s ability to parse and interpret complex temporal acoustic features, translating visual representations of audio data into accurate categorical insights.

\subsection{IMU Data Analysis}

\textbf{Context}:
For the IMU data analysis, the model analyzed a plotted graph displaying time-series data from various IMU sensors, including accelerometers and gyroscopes.

\textbf{Input}:
A multi-line graph is shown, where each line represents different aspects of IMU data such as acceleration and rotational motion over time.

\textbf{Task}:
The model is to match the pattern displayed in the IMU data to a possible physical activity from choices like "Running," "Jumping," "Knocking," and "Dancing."

\textbf{Model Output}:
"Knocking" is selected by the multisensory multi-task language model as the activity that best corresponds to the IMU data presented.

\textbf{Rationale}:
The model interprets the regular pattern of sharp peaks at consistent intervals as indicative of knocking. It notes the repetitive nature of the motion and the varying force, typical for an action like knocking on a surface, which would generate a rhythmic and forceful pattern in IMU sensors.

These examples underscore the model's proficiency in interpreting and classifying data from different IoT modalities based on their temporal and sensory characteristics. The ability to analyze such heterogeneous data effectively is crucial for a variety of applications across smart home systems, healthcare monitoring, and industrial automation. Additionally, these dialog examples highlight the model's potential to bridge the gap between raw sensor outputs and actionable insights, empowering users to engage with complex IoT systems more intuitively and effectively.
Each scenario described showcases not only the model's analytical strength but also its potential to transform raw, often opaque sensor data into comprehensible and actionable information. This capability enhances the usability of IoT systems, making them more accessible and beneficial for a wider range of users and applications.

\section{Dataset Documentation \& Intended Uses}\label{sec: data_doc}

In this section, we outline comprehensive documentation and intended uses for our dataset to ensure transparency, accountability, and ease of access for researchers interested in utilizing this benchmark.

\subsection{Dataset Documentation}
To foster clarity and responsible usage, we adhere to several established documentation frameworks:
\begin{itemize}
    \item \textbf{Datasheets for Datasets:} We provide a detailed datasheet that includes the dataset’s motivation, composition, collection process, recommended uses, and limitations. This aims to ensure that all potential users have a clear understanding of how the dataset was created and its intended scope of application.
    \item \textbf{Dataset Nutrition Labels:} Similar to nutrition labels on food products, our dataset nutrition label offers a concise summary of the data contents, including data types, instances count, and a profile of typical data points.
    \item \textbf{Data Statements for NLP:} For components of our dataset applicable to NLP, we include data statements detailing linguistic demographics and speaker information, ensuring transparency in the representation within the data.
    \item \textbf{Data Cards:} Each modality within the dataset is accompanied by a data card that outlines specific characteristics, use cases, and processing procedures, which helps in understanding each part of the dataset holistically.
    \item \textbf{Accountability Frameworks:} Our dataset complies with existing accountability frameworks to ensure ethical use and application, including guidelines for addressing potential biases and misuse.
\end{itemize}

\input{SECTIONS/Appendix/datasheet}

\subsection{Access and Download Links}
\textbf{Dataset Access:}
The dataset can be viewed and downloaded via our dedicated website, accessible through the following URL: \url{https://github.com/Multi-IoT/MultiIoT}. 
This website provides structured access to the data, along with visualization tools and download options.

\textbf{Metadata Record:}
A comprehensive metadata record is available through our Croissant metadata entry, which can be viewed and downloaded using the following link: \url{https://github.com/Multi-IoT/MultiIoT}. 
This metadata follows the structure suggested by the MLCommons Croissant project, ensuring standardized documentation.

\subsection{Legal and Ethical Assurance}
\textbf{Author Responsibility:}
The authors bear full responsibility for the dataset and confirm that all data was collected and distributed in compliance with all applicable laws and regulations. The dataset does not violate any rights or privacy of individuals or entities.

\textbf{Data Licensing:}
The dataset is released under the Creative Commons Attribution 4.0 International License.

\subsection{Hosting, Licensing, and Maintenance}
\textbf{Hosting Platform:}
The dataset is hosted on our website, which ensures reliable and scalable access to the data. 

\textbf{Maintenance Plan:}
We commit to maintaining the dataset with regular updates and corrections as necessary. The maintenance log will be publicly available to ensure transparency in how the dataset evolves over time.

\textbf{Community Engagement:}
We encourage the community to contribute to the dataset’s improvement by submitting issues or suggestions through our repository’s issue tracker on GitHub.

%% file: SECTIONS/Appendix/vis_imu.tex
\begin{figure}[t]
\centering
\includegraphics[width=0.98\linewidth]{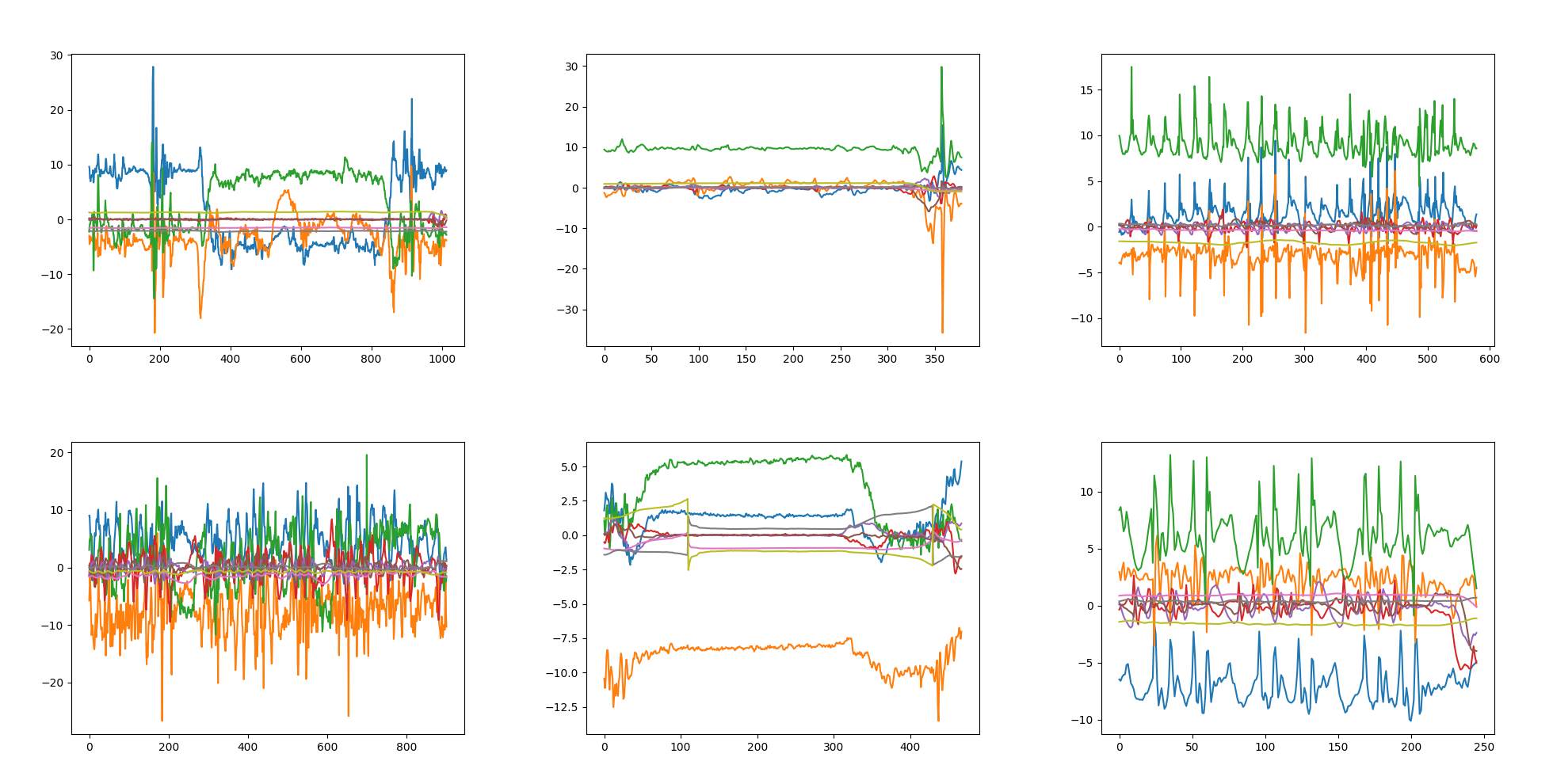}
\vspace{-4mm}
\caption{IMU Visualizations}
\vspace{-4mm}
\label{fig: vis_imu}
\end{figure}

%% file: SECTIONS/Appendix/vis_audio.tex
\begin{figure}[t]
\centering
\includegraphics[width=0.98\linewidth]{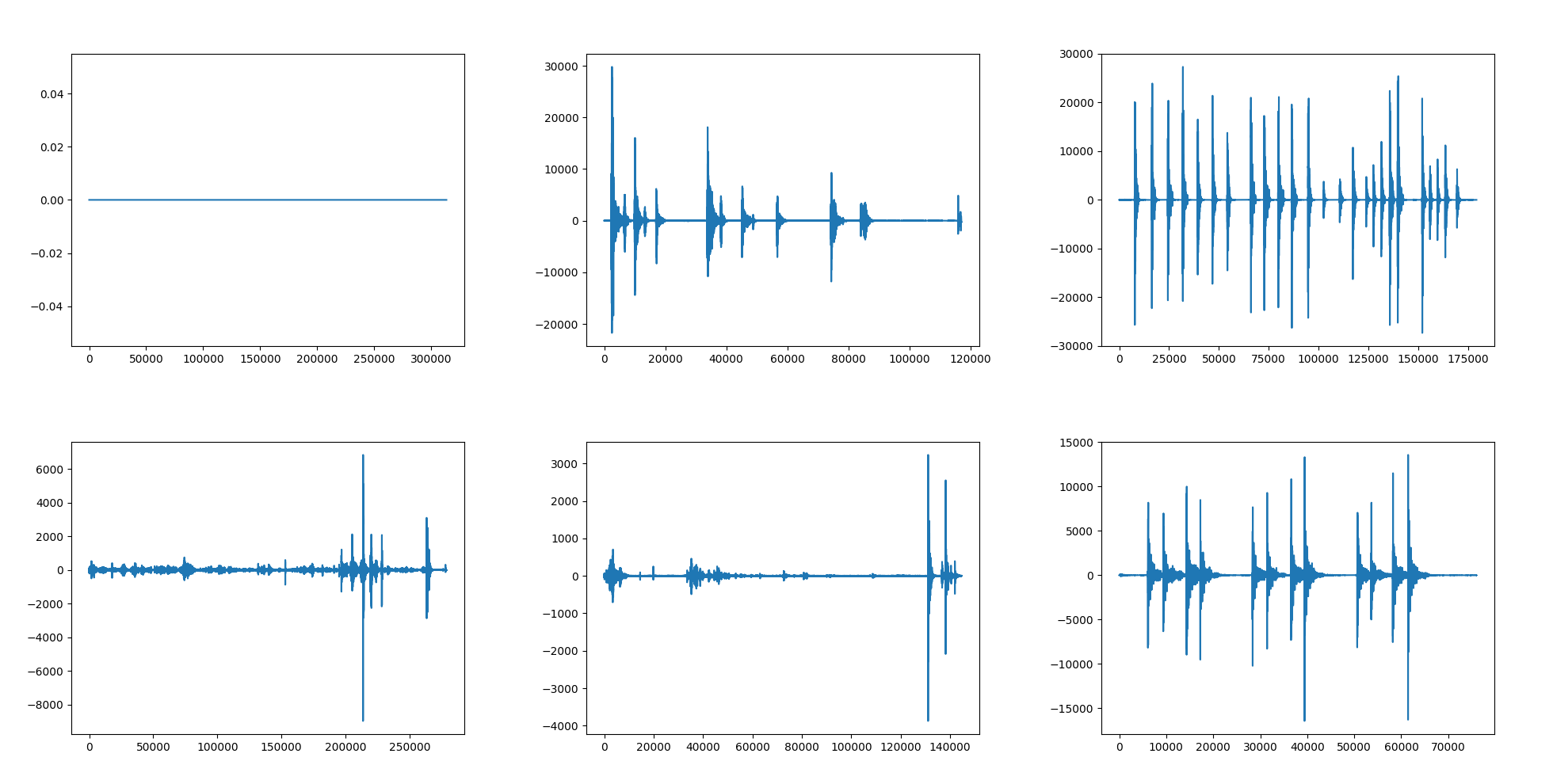}
\vspace{-4mm}
\caption{Audio Visualizations}
\vspace{-4mm}
\label{fig: vis_audio}
\end{figure}

%% file: SECTIONS/Appendix/vis_cap.tex
\begin{figure}[t]
\centering
\includegraphics[width=0.98\linewidth]{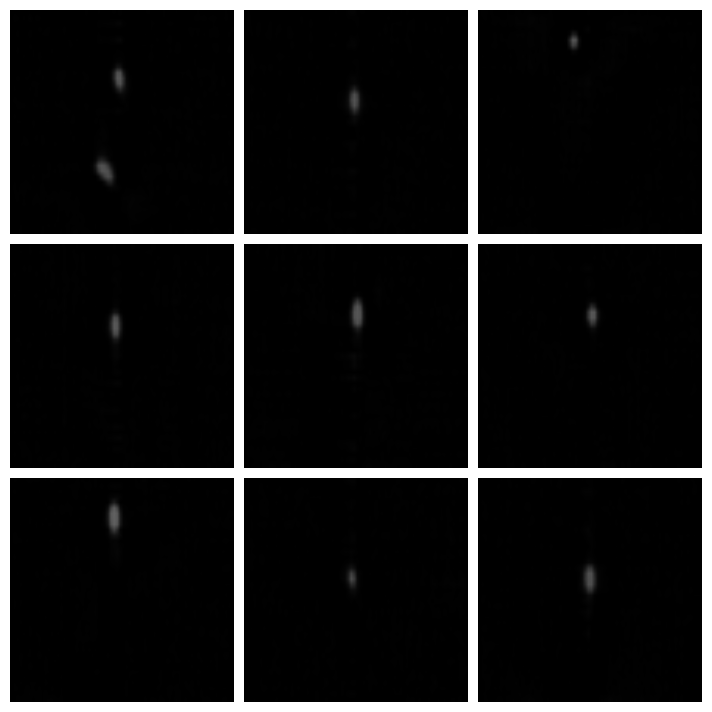}
\vspace{-4mm}
\caption{Capacitance Visualizations}
\vspace{-4mm}
\label{fig: vis_cap}
\end{figure}

%% file: SECTIONS/Appendix/vis_depth.tex
\begin{figure}[t]
\centering
\includegraphics[width=0.98\linewidth]{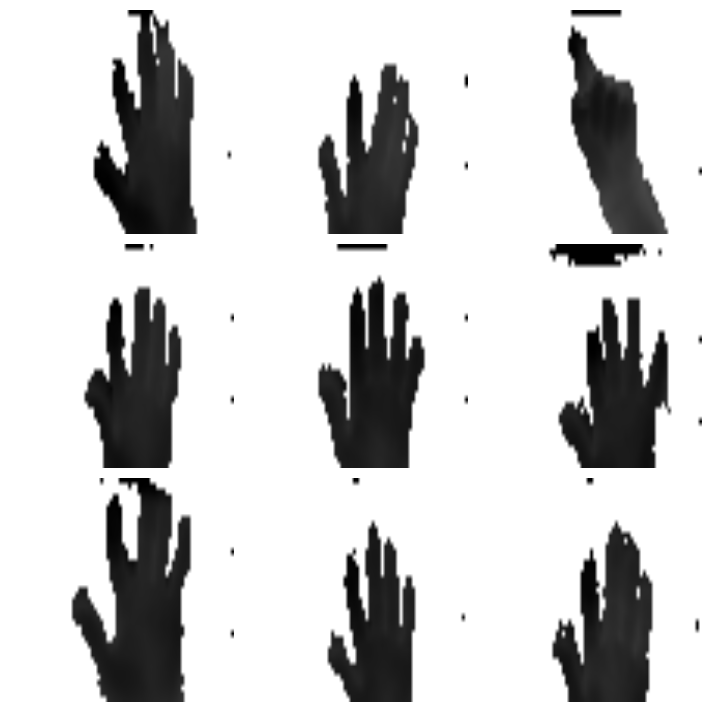}
\vspace{-4mm}
\caption{Depth Visualizations}
\vspace{-4mm}
\label{fig: vis_depth}
\end{figure}

%% file: SECTIONS/Appendix/vis_dialog.tex
\begin{figure}[t]
\centering
\includegraphics[width=0.98\linewidth]{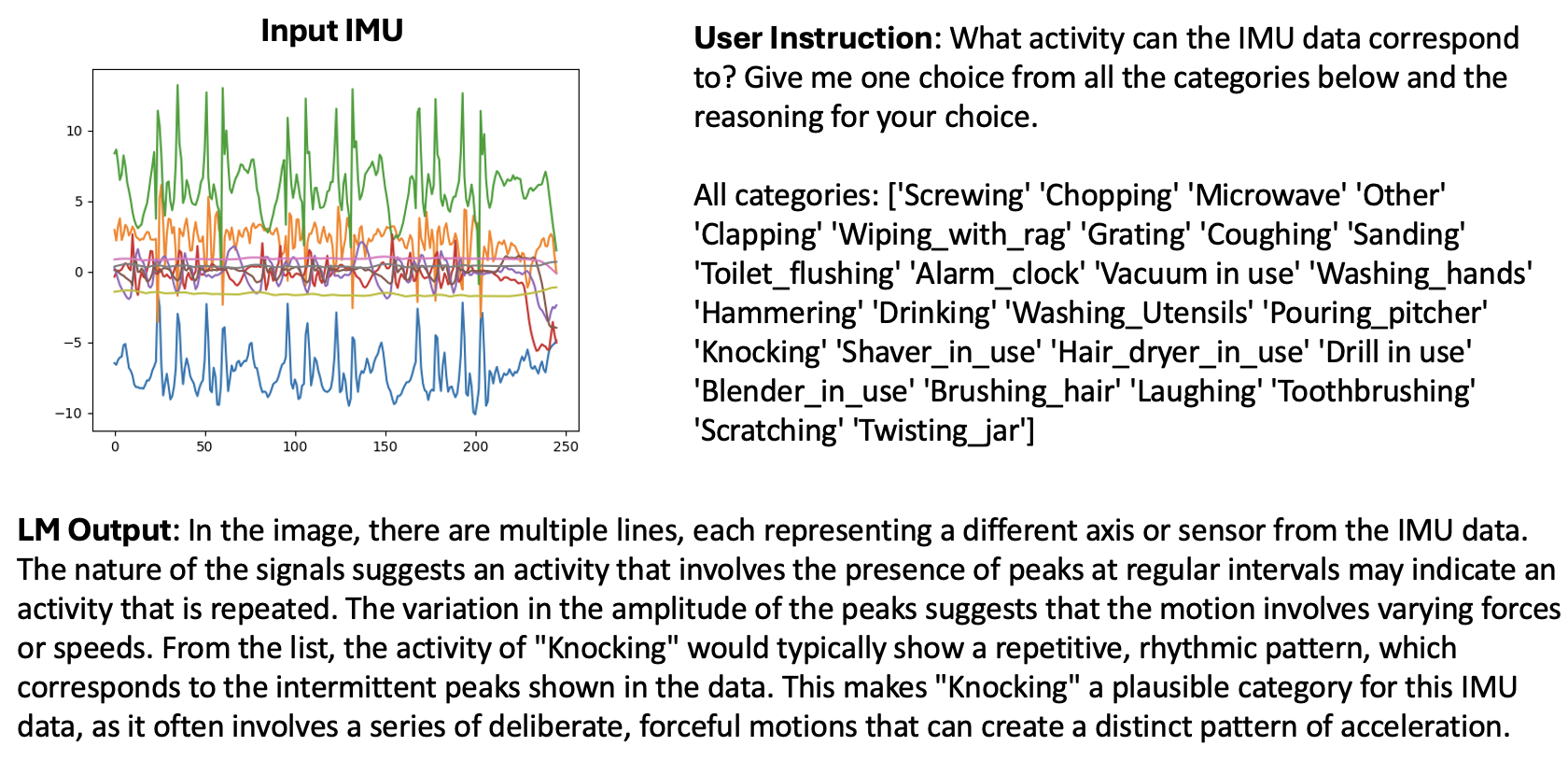}
\vspace{-4mm}
\caption{Dialog for IMU example.}
\vspace{-4mm}
\label{fig: vis_dialog_audio}
\end{figure}

\begin{figure}[t]
\centering
\includegraphics[width=0.98\linewidth]{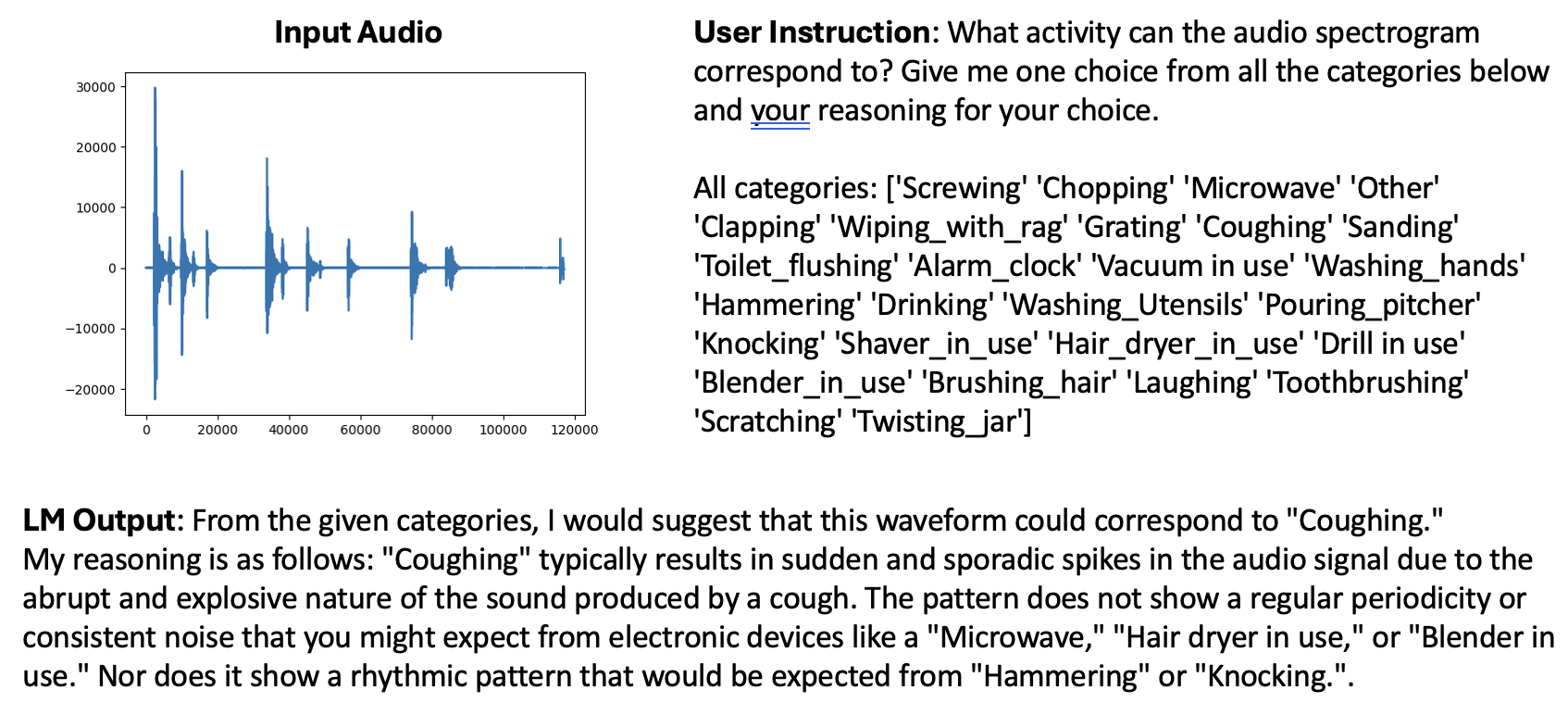}
\vspace{-4mm}
\caption{Dialog for audio example.}
\vspace{-4mm}
\label{fig: vis_dialog_audio}
\end{figure}

%% file: SECTIONS/Appendix/datasheet.tex
\noindent
\textbf{
This document is based on \textit{Datasheets for Datasets} by Gebru \textit{et
al.}~\cite{gebru2018datasheets}. 
}

\begin{mdframed}[linecolor=\sectioncolor]
\section*{\textcolor{\sectioncolor}{
    MOTIVATION
}}
\end{mdframed}

    \textcolor{\sectioncolor}{\textbf{
    For what purpose was the dataset created?
    }
    Was there a specific task in mind? Was there
    a specific gap that needed to be filled? Please provide a description.
    } \\
    The dataset was created to address the lack of large-scale, diverse, multimodal datasets that can be used to improve and evaluate IoT and machine learning models' ability to interpret and process multimodal data streams in real-world scenarios.  \\
    
    \textcolor{\sectioncolor}{\textbf{
    Who created this dataset (e.g., which team, research group) and on behalf
    of which entity (e.g., company, institution, organization)?
    }
    } \\
    This dataset was created by the authors. \\
    
    \textcolor{\sectioncolor}{\textbf{
    What support was needed to make this dataset?
    }
    (e.g.who funded the creation of the dataset? If there is an associated
    grant, provide the name of the grantor and the grant name and number, or if
    it was supported by a company or government agency, give those details.)
    } \\
    No. This dataset was not supported by any grants from several research funding agencies. \\
    
    \textcolor{\sectioncolor}{\textbf{
    Any other comments?
    }} \\
    No. \\

\begin{mdframed}[linecolor=\sectioncolor]
\section*{\textcolor{\sectioncolor}{
    COMPOSITION
}}
\end{mdframed}
    \textcolor{\sectioncolor}{\textbf{
    What do the instances that comprise the dataset represent (e.g., documents,
    photos, people, countries)?
    }
    Are there multiple types of instances (e.g., movies, users, and ratings;
    people and interactions between them; nodes and edges)? Please provide a
    description.
    } \\
    The instances represent a combination of sensor data including audio, visual (image and video), and temporal sensor data (IMU, GPS). \\
    
    \textcolor{\sectioncolor}{\textbf{
    How many instances are there in total (of each type, if appropriate)?
    }
    } \\
    The dataset contains over 1.15 million instances.\\
    
    \textcolor{\sectioncolor}{\textbf{
    Does the dataset contain all possible instances or is it a sample (not
    necessarily random) of instances from a larger set?
    }
    If the dataset is a sample, then what is the larger set? Is the sample
    representative of the larger set (e.g., geographic coverage)? If so, please
    describe how this representativeness was validated/verified. If it is not
    representative of the larger set, please describe why not (e.g., to cover a
    more diverse range of instances, because instances were withheld or
    unavailable).
    } \\
    Each instance consists of raw sensor data along with processed features, including extracted metadata and precomputed sensory features. \\
    
    \textcolor{\sectioncolor}{\textbf{
    What data does each instance consist of?
    }
    “Raw” data (e.g., unprocessed text or images) or features? In either case,
    please provide a description.
    } \\
    Yes, each instance is labeled with activity tags, environmental context, and temporal markers where applicable. \\
    
    \textcolor{\sectioncolor}{\textbf{
    Is there a label or target associated with each instance?
    }
    If so, please provide a description.
    } \\
    Yes, relationships such as sequential and contextual linkages are explicitly defined, enabling the study of interactions across time and modality. \\
    
    \textcolor{\sectioncolor}{\textbf{
    Is any information missing from individual instances?
    }
    If so, please provide a description, explaining why this information is
    missing (e.g., because it was unavailable). This does not include
    intentionally removed information, but might include, e.g., redacted text.
    } \\
    No. \\
    
    \textcolor{\sectioncolor}{\textbf{
    Are relationships between individual instances made explicit (e.g., users’
    movie ratings, social network links)?
    }
    If so, please describe how these relationships are made explicit.
    } \\
    No. \\
    
    \textcolor{\sectioncolor}{\textbf{
    Are there recommended data splits (e.g., training, development/validation,
    testing)?
    }
    If so, please provide a description of these splits, explaining the
    rationale behind them.
    } \\
    Yes. The dataset is split into training (70\%), validation (15\%), and testing (15\%) sets, designed to ensure comprehensive coverage of various scenarios and conditions in each split. \\
    
    \textcolor{\sectioncolor}{\textbf{
    Are there any errors, sources of noise, or redundancies in the dataset?
    }
    If so, please provide a description.
    } \\
    No. \\
    
    \textcolor{\sectioncolor}{\textbf{
    Is the dataset self-contained, or does it link to or otherwise rely on
    external resources (e.g., websites, tweets, other datasets)?
    }
    If it links to or relies on external resources, a) are there guarantees
    that they will exist, and remain constant, over time; b) are there official
    archival versions of the complete dataset (i.e., including the external
    resources as they existed at the time the dataset was created); c) are
    there any restrictions (e.g., licenses, fees) associated with any of the
    external resources that might apply to a future user? Please provide
    descriptions of all external resources and any restrictions associated with
    them, as well as links or other access points, as appropriate.
    } \\
    No. \\
    
    \textcolor{\sectioncolor}{\textbf{
    Does the dataset contain data that might be considered confidential (e.g.,
    data that is protected by legal privilege or by doctor-patient
    confidentiality, data that includes the content of individuals’ non-public
    communications)?
    }
    If so, please provide a description.
    } \\
     No. \\
    
    \textcolor{\sectioncolor}{\textbf{
    Does the dataset contain data that, if viewed directly, might be offensive,
    insulting, threatening, or might otherwise cause anxiety?
    }
    If so, please describe why.
    } \\
     No. \\
    
    \textcolor{\sectioncolor}{\textbf{
    Does the dataset relate to people?
    }
    If not, you may skip the remaining questions in this section.
    } \\
     No. \\
    
    \textcolor{\sectioncolor}{\textbf{
    Does the dataset identify any subpopulations (e.g., by age, gender)?
    }
    If so, please describe how these subpopulations are identified and
    provide a description of their respective distributions within the dataset.
    } \\
     No. \\
    
    \textcolor{\sectioncolor}{\textbf{
    Is it possible to identify individuals (i.e., one or more natural persons),
    either directly or indirectly (i.e., in combination with other data) from
    the dataset?
    }
    If so, please describe how.
    } \\
     No. \\
    
    \textcolor{\sectioncolor}{\textbf{
    Does the dataset contain data that might be considered sensitive in any way
    (e.g., data that reveals racial or ethnic origins, sexual orientations,
    religious beliefs, political opinions or union memberships, or locations;
    financial or health data; biometric or genetic data; forms of government
    identification, such as social security numbers; criminal history)?
    }
    If so, please provide a description.
    } \\
    No. \\
    
    \textcolor{\sectioncolor}{\textbf{
    Any other comments?
    }} \\
     No. \\

\begin{mdframed}[linecolor=\sectioncolor]
\section*{\textcolor{\sectioncolor}{
    COLLECTION
}}
\end{mdframed}

    \textcolor{\sectioncolor}{\textbf{
    How was the data associated with each instance acquired?
    }
    Was the data directly observable (e.g., raw text, movie ratings),
    reported by subjects (e.g., survey responses), or indirectly
    inferred/derived from other data (e.g., part-of-speech tags, model-based
    guesses for age or language)? If data was reported by subjects or
    indirectly inferred/derived from other data, was the data
    validated/verified? If so, please describe how.
    } \\
    No. \\
    
    \textcolor{\sectioncolor}{\textbf{
    Over what timeframe was the data collected?
    }
    Does this timeframe match the creation timeframe of the data associated
    with the instances (e.g., recent crawl of old news articles)? If not,
    please describe the timeframe in which the data associated with the
    instances was created. Finally, list when the dataset was first published.
    } \\
    Data collection spanned over half one year. \\
    
    \textcolor{\sectioncolor}{\textbf{
    What mechanisms or procedures were used to collect the data (e.g., hardware
    apparatus or sensor, manual human curation, software program, software
    API)?
    }
    How were these mechanisms or procedures validated?
    } \\
    Different sensors were used to collect the sensory data. \\
    
    \textcolor{\sectioncolor}{\textbf{
    What was the resource cost of collecting the data?
    }
    (e.g. what were the required computational resources, and the associated
    financial costs, and energy consumption - estimate the carbon footprint.
    See Strubell \textit{et al.}\cite{strubell2019energy} for approaches in this area.)
    } \\
    We use V100 \& A100 GPUs to curate data and train our models. \\
    
    \textcolor{\sectioncolor}{\textbf{
    If the dataset is a sample from a larger set, what was the sampling
    strategy (e.g., deterministic, probabilistic with specific sampling
    probabilities)?
    }
    } \\
    No. The dataset is not a subset of a larger set. \\
    
    \textcolor{\sectioncolor}{\textbf{
    Who was involved in the data collection process (e.g., students,
    crowdworkers, contractors) and how were they compensated (e.g., how much
    were crowdworkers paid)?
    }
    } \\
    Authors are involved in the data curation process. \\
    
    \textcolor{\sectioncolor}{\textbf{
    Were any ethical review processes conducted (e.g., by an institutional
    review board)?
    }
    If so, please provide a description of these review processes, including
    the outcomes, as well as a link or other access point to any supporting
    documentation.
    } \\
    No. \\
    
    \textcolor{\sectioncolor}{\textbf{
    Does the dataset relate to people?
    }
    If not, you may skip the remainder of the questions in this section.
    } \\
    No. \\
    
    \textcolor{\sectioncolor}{\textbf{
    Did you collect the data from the individuals in question directly, or
    obtain it via third parties or other sources (e.g., websites)?
    }
    } \\
    No. \\
    
    \textcolor{\sectioncolor}{\textbf{
    Were the individuals in question notified about the data collection?
    }
    If so, please describe (or show with screenshots or other information) how
    notice was provided, and provide a link or other access point to, or
    otherwise reproduce, the exact language of the notification itself.
    } \\
    No. \\
    
    \textcolor{\sectioncolor}{\textbf{
    Did the individuals in question consent to the collection and use of their
    data?
    }
    If so, please describe (or show with screenshots or other information) how
    consent was requested and provided, and provide a link or other access
    point to, or otherwise reproduce, the exact language to which the
    individuals consented.
    } \\
    No. \\
    
    \textcolor{\sectioncolor}{\textbf{
    If consent was obtained, were the consenting individuals provided with a
    mechanism to revoke their consent in the future or for certain uses?
    }
     If so, please provide a description, as well as a link or other access
     point to the mechanism (if appropriate)
    } \\
    No. \\
    
    \textcolor{\sectioncolor}{\textbf{
    Has an analysis of the potential impact of the dataset and its use on data
    subjects (e.g., a data protection impact analysis)been conducted?
    }
    If so, please provide a description of this analysis, including the
    outcomes, as well as a link or other access point to any supporting
    documentation.
    } \\
    No. \\
    
    \textcolor{\sectioncolor}{\textbf{
    Any other comments?
    }} \\
    No. \\

\begin{mdframed}[linecolor=\sectioncolor]
\section*{\textcolor{\sectioncolor}{
    PREPROCESSING / CLEANING / LABELING
}}
\end{mdframed}

    \textcolor{\sectioncolor}{\textbf{
    Was any preprocessing/cleaning/labeling of the data
    done(e.g.,discretization or bucketing, tokenization, part-of-speech
    tagging, SIFT feature extraction, removal of instances, processing of
    missing values)?
    }
    If so, please provide a description. If not, you may skip the remainder of
    the questions in this section.
    } \\
    No. \\

    \textcolor{\sectioncolor}{\textbf{
    Was the “raw” data saved in addition to the preprocessed/cleaned/labeled
    data (e.g., to support unanticipated future uses)?
    }
    If so, please provide a link or other access point to the “raw” data.
    } \\
    No. \\

    \textcolor{\sectioncolor}{\textbf{
    Is the software used to preprocess/clean/label the instances available?
    }
    If so, please provide a link or other access point.
    } \\
    No. \\

    \textcolor{\sectioncolor}{\textbf{
    Any other comments?
    }} \\
    No. \\

\begin{mdframed}[linecolor=\sectioncolor]
\section*{\textcolor{\sectioncolor}{
    USES
}}
\end{mdframed}

    \textcolor{\sectioncolor}{\textbf{
    Has the dataset been used for any tasks already?
    }
    If so, please provide a description.
    } \\
    No. \\

    \textcolor{\sectioncolor}{\textbf{
    Is there a repository that links to any or all papers or systems that use the dataset?
    }
    If so, please provide a link or other access point.
    } \\
    No. \\

    \textcolor{\sectioncolor}{\textbf{
    What (other) tasks could the dataset be used for?
    }
    } \\
    Beyond the current uses, the dataset holds potential for tasks in real-world IoT applications. \\

    \textcolor{\sectioncolor}{\textbf{
    Is there anything about the composition of the dataset or the way it was
    collected and preprocessed/cleaned/labeled that might impact future uses?
    }
    For example, is there anything that a future user might need to know to
    avoid uses that could result in unfair treatment of individuals or groups
    (e.g., stereotyping, quality of service issues) or other undesirable harms
    (e.g., financial harms, legal risks) If so, please provide a description.
    Is there anything a future user could do to mitigate these undesirable
    harms?
    } \\
    No. \\

    \textcolor{\sectioncolor}{\textbf{
    Are there tasks for which the dataset should not be used?
    }
    If so, please provide a description.
    } \\
    No. \\

    \textcolor{\sectioncolor}{\textbf{
    Any other comments?
    }} \\
    No. \\

\begin{mdframed}[linecolor=\sectioncolor]
\section*{\textcolor{\sectioncolor}{
    DISTRIBUTION
}}
\end{mdframed}

    \textcolor{\sectioncolor}{\textbf{
    Will the dataset be distributed to third parties outside of the entity
    (e.g., company, institution, organization) on behalf of which the dataset
    was created?
    }
    If so, please provide a description.
    } \\
    No. \\

    \textcolor{\sectioncolor}{\textbf{
    How will the dataset will be distributed (e.g., tarball on website, API,
    GitHub)?
    }
    Does the dataset have a digital object identifier (DOI)?
    } \\
    The dataset is available for download via a website page. \\

    \textcolor{\sectioncolor}{\textbf{
    When will the dataset be distributed?
    }
    } \\
    The dataset will be available upon publication. \\

    \textcolor{\sectioncolor}{\textbf{
    Will the dataset be distributed under a copyright or other intellectual
    property (IP) license, and/or under applicable terms of use (ToU)?
    }
    If so, please describe this license and/or ToU, and provide a link or other
    access point to, or otherwise reproduce, any relevant licensing terms or
    ToU, as well as any fees associated with these restrictions.
    } \\
    No. \\

    \textcolor{\sectioncolor}{\textbf{
    Have any third parties imposed IP-based or other restrictions on the data
    associated with the instances?
    }
    If so, please describe these restrictions, and provide a link or other
    access point to, or otherwise reproduce, any relevant licensing terms, as
    well as any fees associated with these restrictions.
    } \\
    No. \\

    \textcolor{\sectioncolor}{\textbf{
    Do any export controls or other regulatory restrictions apply to the
    dataset or to individual instances?
    }
    If so, please describe these restrictions, and provide a link or other
    access point to, or otherwise reproduce, any supporting documentation.
    } \\
    No. \\

    \textcolor{\sectioncolor}{\textbf{
    Any other comments?
    }} \\
    No. \\

\begin{mdframed}[linecolor=\sectioncolor]
\section*{\textcolor{\sectioncolor}{
    MAINTENANCE
}}
\end{mdframed}

    \textcolor{\sectioncolor}{\textbf{
    Who is supporting/hosting/maintaining the dataset?
    }
    } \\
    The dataset is maintained by the authors. \\

    \textcolor{\sectioncolor}{\textbf{
    How can the owner/curator/manager of the dataset be contacted (e.g., email
    address)?
    }
    } \\
    The owner of the dataset can contacted by email. \\

    \textcolor{\sectioncolor}{\textbf{
    Is there an erratum?
    }
    If so, please provide a link or other access point.
    } \\
    No. \\

    \textcolor{\sectioncolor}{\textbf{
    Will the dataset be updated (e.g., to correct labeling errors, add new
    instances, delete instances)?
    }
    If so, please describe how often, by whom, and how updates will be
    communicated to users (e.g., mailing list, GitHub)?
    } \\
    No. \\

    \textcolor{\sectioncolor}{\textbf{
    If the dataset relates to people, are there applicable limits on the
    retention of the data associated with the instances (e.g., were individuals
    in question told that their data would be retained for a fixed period of
    time and then deleted)?
    }
    If so, please describe these limits and explain how they will be enforced.
    } \\
    No. \\

    \textcolor{\sectioncolor}{\textbf{
    Will older versions of the dataset continue to be
    supported/hosted/maintained?
    }
    If so, please describe how. If not, please describe how its obsolescence
    will be communicated to users.
    } \\
    Yes. Older versions will be archived and accessible for historical comparison and research consistency. \\

    \textcolor{\sectioncolor}{\textbf{
    If others want to extend/augment/build on/contribute to the dataset, is
    there a mechanism for them to do so?
    }
    If so, please provide a description. Will these contributions be
    validated/verified? If so, please describe how. If not, why not? Is there a
    process for communicating/distributing these contributions to other users?
    If so, please provide a description.
    } \\
    Yes. Feedback and contributions from the community are highly encouraged and can be facilitated through our repository.

    \textcolor{\sectioncolor}{\textbf{
    Any other comments?
    }} \\
    No. \\